\documentclass[pdflatex,sn-nature]{sn-jnl}

\usepackage{graphicx}%
\usepackage{multirow}%
\usepackage{amsmath,amssymb,amsfonts}%
\usepackage{amsthm}%
\usepackage{mathrsfs}%
\usepackage[title]{appendix}%
\usepackage{xcolor}%
\usepackage{textcomp}%
\usepackage{manyfoot}%
\usepackage{booktabs}%
\usepackage{algorithm}%
\usepackage{algorithmicx}%
\usepackage{algpseudocode}%
\usepackage{listings}%
\usepackage{hyperref}
\usepackage{ragged2e}
\usepackage{float}
\usepackage{placeins}

\theoremstyle{thmstyleone}%
\theoremstyle{thmstyletwo}%

\theoremstyle{thmstylethree}%

\begin{document}

\title[Article Title]{Creating and Evaluating Code-Mixed Nepali-English and Telugu-English Datasets for Abusive Language Detection Using Traditional and Deep Learning Models}


\author[1]{\fnm{Manish} \sur{Pandey}}\email{manish\_pandey@srmap.edu.in}

\author[1]{\fnm{Nageshwar Prasad} \sur{Yadav}}\email{nageshwar\_prasadyadav@srmap.edu.in}

\author[1]{\fnm{Mokshada} \sur{Adduru}}\email{mokshada\_adduru@srmap.edu.in}

\author*[1]{\fnm{Sawan} \sur{Rai}}\email{sawan.r@srmap.edu.in}

\affil*[1]{\orgdiv{Dept. of Computer Science and Engineering}, \orgname{SRM University AP}, \orgaddress{ \city{Guntur}, \postcode{522502}, \state{AP}, \country{India}}}


\abstract{With the growing presence of multilingual users on social media, detecting abusive language in code-mixed text has become increasingly challenging. Code-mixed communication, where users seamlessly switch between English and their native languages, poses difficulties for traditional abuse detection models, as offensive content may be context-dependent or obscured by linguistic blending. While abusive language detection has been extensively explored for high-resource languages like English and Hindi, low-resource languages such as Telugu and Nepali remain underrepresented, leaving gaps in effective moderation.
In this study, we introduce a novel, manually annotated dataset of 2 thousand Telugu-English and 5 Nepali-English code-mixed comments, categorized as abusive and non-abusive, collected from various social media platforms. The dataset undergoes rigorous preprocessing before being evaluated across multiple Machine Learning (ML), Deep Learning (DL), and Large Language Models (LLMs). We experimented with models including Logistic Regression, Random Forest, Support Vector Machines (SVM), Neural Networks (NN), LSTM, CNN, and LLMs, optimizing their performance through hyperparameter tuning, and evaluate it using 10-fold cross-validation and statistical significance testing (t-test).
Our findings provide key insights into the challenges of detecting abusive language in code-mixed settings and offer a comparative analysis of computational approaches. This study contributes to advancing NLP for low-resource languages by establishing benchmarks for abusive language detection in Telugu-English and Nepali-English code-mixed text. The dataset and insights can aid in the development of more robust moderation strategies for multilingual social media environments.}




\maketitle

\section{Introduction}
Social media has become an important space where people share opinions, connect with others, and express themselves freely. However, with its growing popularity, there has been an increase in online abuse and hate speech, making these platforms more toxic and unsafe for many users. One major challenge in detecting abusive content—especially in countries like India and Nepal—is the use of code-mixing, where users switch between languages like Telugu or Nepali and English within the same sentence or post. This form of multilingual communication feels natural to many speakers, but it creates difficulties for automated abuse detection systems\cite{srivastava2020phinc}.

Most tools designed to detect hate speech or offensive language are trained on English text, and they often struggle when confronted with code-mixed content \cite{huzaifah2024evaluating}. This is because code-mixed text is messy—it includes inconsistent spelling, grammar from more than one language, and Romanized native words, which aren’t always spelled consistently. For instance, someone might write “nuvvu chesina thing chala worst,” mixing Romanized Telugu with English. Similarly, in Nepali-English, a user might say “timro attitude dherai irritating xa,” combining Nepali syntax and vocabulary with English words. A system trained only in English would fail to understand the full meaning or even recognize that this could be abusive. These challenges make it difficult for traditional models to accurately classify such content.

Moreover, there is a lack of properly labeled datasets for code-mixed languages. Telugu-English and Nepali-English, in particular, lack sufficient labeled data for training and testing abuse detection models \cite{singh2018twitter}, \cite{mandl2019overview}. Most available datasets are small, unbalanced, or not annotated for abusive content at all. Without enough high-quality examples, building effective models becomes difficult. Additionally, because code-mixed language reflects cultural and social context, detecting abuse becomes even more complex—it’s not only about the words used but also about when and how they are used.

Another unique challenge with code-mixed text is the subjectivity of meaning in different cultural and regional contexts as discussed by Authors in \cite{tiwari2021mind}. A phrase that may seem neutral to someone unfamiliar with the culture could carry deeply offensive or sarcastic undertones in certain communities. Abusers often exploit these linguistic grey areas to bypass content moderation systems. This underscores the importance of training models not just on code-mixed data, but on data that reflects the nuanced language patterns, tone, and slang commonly used in Telugu-English or Nepali-English online interactions \cite{ruder2019survey}.

Furthermore, transliteration adds another layer of complexity. Since Telugu and Nepali are written in their native scripts (Telugu and Devanagari, respectively), users often type them in Roman script on social media platforms. However, there are no strict rules for transliterating these words \cite{barman2014code}. The same word can be spelled differently by different users—for example, “chepparu,” “cheparu,” or “cheppaaru” could all mean the same thing in Telugu. Similarly, “kasto” in Nepali might appear as “kasto,” “kasto,” or even “casto,” depending on the user. These inconsistencies confuse models that haven’t been trained on a wide variety of such forms, making basic text preprocessing more challenging.

Despite the growing number of multilingual users online, limited focus has been given to building datasets and tools for low-resource code-mixed languages. Most research has centered around more widely studied combinations like Hindi-English or Spanish-English \cite{solorio2014overview}. Languages like Telugu and Nepali, on the other hand, receive relatively little attention in NLP research, despite being spoken by millions. This creates a gap in resources and tools available to these communities, reflecting a kind of digital inequality—where some language groups are better protected from online abuse simply because they are better represented in research and development.

To address this issue, our research introduces a manually annotated dataset containing 5,000 Nepali-English and 2,000 Telugu-English code-mixed abusive comments. We focused on collecting realistic, social media-style examples and carefully labeled them for accuracy. We then evaluated several machine learning and deep learning models—including Logistic Regression, SVM, Random Forest, Neural Networks, LSTM, and CNN—to determine which performed best. The models were tested using 10-fold cross-validation, and we also used statistical tests (such as T-tests) to fairly compare their performance.

This paper not only fills a gap in available data but also provides insight into how different models handle code-mixed abuse detection. Our findings demonstrate that, with the right dataset and proper evaluation, it is possible to build better tools for moderating content in multilingual online communities. This is particularly important for making the internet safer and more inclusive for speakers of low-resource languages like Telugu and Nepali, who are often overlooked in mainstream AI and NLP research.


Furthermore, the findings emphasize the importance of not only relying on automated systems but also ensuring that the data used for training reflects the language spoken by the target community. A more culturally and linguistically diverse dataset, along with improved preprocessing techniques to handle transliteration variations and mixed-language syntax, could lead to more robust models for detecting abusive content. By addressing these challenges, we can work toward creating a safer online environment for users who communicate in code-mixed languages like Telugu and Nepali.\\

The rest of this paper is organized as follows. 
Section~\ref{sec:code_mixing} discusses the notion of code-mixing within the context of Nepali and Telugu languages, giving linguistic and sociocultural background. 
Section~\ref{sec:data_generation} describes the process of generating code-mixed datasets for these languages, including the approaches adopted for data collection, sources of data, preprocessing, annotation strategies, an overview of the datasets, the difficulties faced during the process, and the ethical considerations upheld. 
Section~\ref{sec:dataset_properties} discusses the properties and characteristics of the curated datasets. 
Section~\ref{sec:experiments} reports experiments using these corpora, after which Section~\ref{sec:results} describes results found. 
Section~\ref{sec:discussion} outlines an in-depth discussion of results, and Section~\ref{sec:related_work} offers an overview of previous work relevant to code-mixing as well as multilingual NLP. 
Section~\ref{sec:conclusion} concludes the paper, including a summary of major contributions along with future prospects.

\section{Code-Mixing in Telugu and Nepali}\label{sec:code_mixing}
Code-mixing, the practice of alternating between two or more languages within a sentence or discourse, is a common phenomenon in multilingual societies. In countries like Nepal and India, where multiple languages coexist, code-mixing has become an essential feature of communication, particularly among younger generations. In these regions, the integration of English with native languages such as Nepali and Telugu is especially prevalent. Several studies have observed this trend across various media and conversational contexts, highlighting how English is increasingly embedded within informal and semi-formal discourse in these societies \cite{adhikari2018english, saud2022linguistic, gurung2019nepali}. This section explores the use of English within Nepali and Telugu, examining its role in modern communication, the contexts in which it occurs, and the patterns of language switching. The impact of code-mixing on language structure, expression, and comprehension will also be discussed.

\subsection{Code-Mixing in Nepali}
Code-mixing between Nepali and English is common in Nepal, especially in urban areas and among younger generations. English is frequently used in education, media, and business, which leads Nepali speakers to blend English words and phrases into their speech. This is particularly evident when discussing modern, technological, or global topics. Code-mixing reflects the influence of global culture and serves as a practical way to switch between languages for ease and efficiency.

Example of Nepali-English Code-Mixing:

Nepali-English: "Timi le test paper complete garne hoina ra? I think you should finish it soon."

Translation: "Aren't you going to complete the test paper? I think you should finish it soon."

In this example, the sentence is in Nepali structure, but English words like "test paper" and "finish" are used for efficiency and because there may be no direct translation for these terms in Nepali.

\subsection{Code-Mixing in Telugu}
Telugu speakers also commonly mix English into their sentences, especially in informal settings or digital communication. Telugu sentence structure and grammar are maintained, but English words are often inserted, particularly when discussing modern, technical, or pop culture topics. This is particularly true in urban environments, where English is widely used in schools, workplaces, and social media.

Example of Telugu-English Code-Mixing:

Telugu-English: "Me intloo camping ke Yantha mandhi waiting?"

Translation: "How many people are waiting for camping at your house?"

This sentence follows Telugu grammar but incorporates English words like "camping" and "waiting", demonstrating how English is blended into everyday Telugu conversation for terms that are commonly used in the global context.

\subsection{Code-Mixing Patterns}
Code-mixing in both Nepali and Telugu generally occurs in two main forms: intrasentential mixing and intersentential mixing.

Intrasentential mixing involves switching languages within a single sentence. For example, parts of the sentence may follow one language's grammar and sentence structure, while other parts incorporate words from another language. This type of mixing is often considered the most complex due to the syntactic constraints it imposes \cite{poplack1980, muysken2000bilingual}.

Inter-sentential mixing occurs when the switch happens at sentence boundaries. One sentence might be entirely in one language, followed by another sentence in a different language. This form is typically observed among speakers with lower proficiency in their second language \cite{poplack1980, ali2019inter}.

Both forms of code-mixing allow speakers to seamlessly navigate between languages, helping them express themselves more effectively. However, this can make language processing more complex, as the language switch may obscure the intended meaning or emotion behind the sentence.

\section{Creating the Telugu-Nepali Code-Mixed Dataset}\label{sec:data_generation}
To develop effective abusive language detection models for low-resource, code-mixed languages, well-curated datasets are imperative. This part explains the formation of Romanized Telugu-English and Nepali-English datasets, i.e., collection, preprocessing, annotation, challenges, and ethics.

\subsection{Data Collection}
To tackle the lack of accessible datasets involving abusive content in code-mixed low-resource languages, this study focused on collecting Romanized Telugu-English and Nepali-English text data from social media. The aim was to build a diverse and realistic dataset to help identify abusive language in naturally occurring code-switched conversations.

\subsection{Data Sources}
Comments were manually gathered from platforms like Twitter, YouTube, and Instagram. These platforms were chosen for their large volumes of user-generated, informal, and code-mixed content. The dataset includes both abusive and non-abusive comments from various topics—politics, entertainment, and everyday life—ensuring a wide range of linguistic tones and styles.

\subsection{Preprocessing Steps}
Prior to analysis, the data went through several systematic cleaning processes to correct quality and consistency. First, all texts were made lowercase to prevent duplication due to case sensitivity. The punctuation marks were stripped off, and tokenization was carried out to divide the sentences into separate words or symbols to facilitate easier processing. Emojis were converted to their textual equivalents, and numerical values were spelled out as words to ensure uniformity of format across the dataset. Special characters and extraneous symbols were also removed, and the cleaned text was then converted to numerical format (vectorized) to facilitate smooth input to machine learning models.

\subsection{Annotation Process}
Comments were manually annotated by individuals fluent in English and either Telugu or Nepali. They were classified into two categories:
\textbf{Abusive}: Containing rude, disrespectful, or offensive language, even if not explicitly promoting hate.
\textbf{Non-abusive}: Free from any offensive, harmful, or inappropriate content.
Annotators were guided to consider tone, cultural context, sarcasm, and linguistic variation. In cases of disagreement, annotations were discussed and finalized collectively to maintain labeling consistency.

\subsection{Dataset Overview}
Telugu-English Dataset contains 2,000 comments in total — 1,200 labeled as abusive and 800 as non-abusive, showing moderate class imbalance. while Nepali-English Dataset includes 5,000 comments — 2,632 abusive and 2,368 non-abusive, offering a more balanced distribution.
These datasets offer valuable resources for building models that address the unique challenges of code-mixed abusive content detection.

\subsection{Challenges Faced}
A key difficulty was the limited availability of Romanized content; most online data exists in native scripts, especially for Telugu and Nepali. Since our focus was on Romanized text, finding relevant content required extensive manual filtering.

Another issue was the inconsistency in Romanization — the same word could be spelled in several ways due to differences in pronunciation or personal habits. This made preprocessing more complex.

The nature of code-mixing added additional challenges. Comments often blended English with native language words, even within a single term. This complicated both the preprocessing and the annotation phases. Sarcasm, slang, and cultural nuances made it harder to categorize comments as abusive or not, often requiring careful judgment.

Lastly, the imbalance in class distribution—especially for Telugu—necessitated focused collection efforts to ensure abusive content was adequately represented without skewing the dataset.

\subsection{Ethical Practices}
All data used was publicly available and collected in line with platform policies, such as Twitter’s Developer Agreement. No personal or sensitive information (like usernames or timestamps) was retained, preserving user privacy. The dataset is solely for academic and research purposes, aimed at contributing to the safe and ethical development of abusive language detection systems. No text was fabricated, altered, or misused during this study.
\FloatBarrier
\section{Analysis of Datasets} \label{sec:dataset_properties}
This section provides an in-depth breakdown of the code-mixed Nepali-English and Telugu-English datasets. It includes data composition, the level of code-mixing using the Code-Mixing Index (CMI), and token distribution language-wise to bring out the linguistic patterns and diversity in the datasets.

\begin{table}[h]
\caption{Data Composition Overview}\label{tab:composition}
\begin{tabular}{@{}lll@{}}
\toprule
\textbf{Feature} & \textbf{Telugu-English} & \textbf{Nepali-English} \\
\midrule
Total Comments & 2,000 & 5,000 \\
Abusive Comments & 1,200 & 2,632 \\
Non-Abusive Comments & 800 & 2,368 \\
Total Tokens & 18,551 & 44,902 \\
Average Tokens per Sentence & 9.28 & 8.98 \\
Domain Diversity & High & High \\
Romanized Language Mix & Telugu + English & Nepali + English \\
Abusive/Non-Abusive Ratio & 60:40 & $\sim$53:47 \\
\bottomrule
\end{tabular}
\end{table}

The Telugu-English dataset consists of 2,000 code-mixed comments. About 60\% are abusive, and the rest are non-abusive. The language used is mostly informal, drawn from social media platforms, showcasing a diverse set of domains like politics, entertainment, and daily life. The comments are written using Roman script and contain tokens from both English and Telugu languages.

The Nepali-English dataset contains 5,000 code-mixed comments with a slightly more balanced 53:47 abusive to non-abusive ratio. The data reflects natural online usage, with Romanized Nepali intermingled with English, especially in technology, current affairs, and youth conversations.

 \subsection{Code-Mixing Statistics}

To measure the extent of code-mixing in the datasets, we calculated the Code-Mixing Index (CMI), a commonly used metric in linguistic analysis. CMI quantifies the level of mixing between multiple languages within a sentence or utterance. It is defined as:

\begin{equation}
CMI = 100 \times \left(1 - \frac{\max(w_i)}{n - u}\right)
\end{equation}

Where:
\begin{itemize}
    \item \( n \) = Total number of tokens in the comment
    \item \( u \) = Number of language-independent tokens (e.g., punctuation, symbols)
    \item \( w_i \) = Number of tokens in language \( i \)
    \item \( \max(w_i) \) = Number of tokens in the most dominant language
\end{itemize}

Higher CMI values indicate greater levels of code-mixing, while a value of 0 corresponds to monolingual content.
This metric was introduced by Gambäck and Das \cite{gamback2014comparing} and has been widely adopted in code-mixed language research \cite{gamback2016comparing}.


The Nepali-English dataset demonstrates a moderate degree of code-mixing. English tokens frequently occur within otherwise Nepali syntactic structures, especially in domains such as pop culture, online discourse, and education. These borrowed words blend into everyday usage, reflecting a natural form of code-switching in digital conversations.

\begin{figure}[H]
    \centering
    \includegraphics[width=0.8\textwidth]{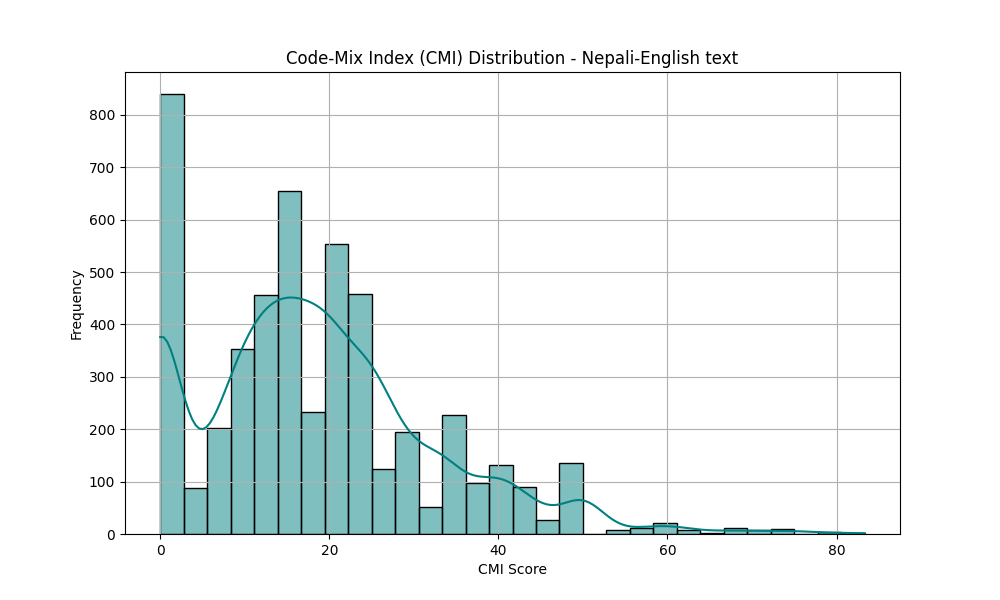}
    \caption{Distribution of Code-Mixing Index (CMI) for Nepali-English dataset.}
    \label{fig:cmi_distribution_nepali}
\end{figure}

Figure 2 shows the distribution of CMI scores in a Telugu-English dataset containing Romanized comments. Most of the comments are concentrated at a CMI score of 0, indicating that they are treated as monolingual. This is likely because Telugu words written in the Roman script are not being identified as Telugu by the current method, causing them to be classified as English. As a result, the true extent of code-mixing in the dataset is likely underrepresented, suggesting a need for better handling of Romanized Telugu in the analysis.
\begin{figure}[H]
    \centering
    \includegraphics[width=0.8\textwidth]{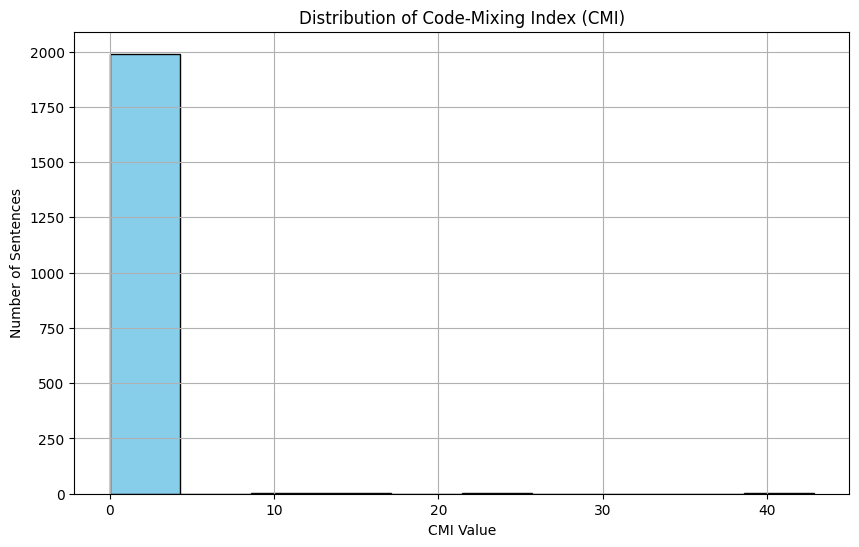}
    \caption{Distribution of Code-Mixing Index (CMI) for Telugu-English dataset.}
    \label{fig:cmi_distribution_telugu}
\end{figure}

\subsection{Language distribution}

This section provides an overview of the distribution of tokens based on their language labels. This analysis helps us understand the syntactic nature and code-mixing dynamics of both the Telugu-English and Nepali-English datasets.

\begin{table}[h]
\caption{Comment-wise Language Distribution in Telugu-English and Nepali-English Datasets}
\label{tab:lang_dist_tel_nep}
\begin{tabular}{@{}lcccc@{}}
\toprule
\multirow{2}{*}{\textbf{Language Category}} & \multicolumn{2}{c}{\textbf{Telugu-English}} & \multicolumn{2}{c}{\textbf{Nepali-English}} \\
\cmidrule(lr){2-3} \cmidrule(lr){4-5}
 & \textbf{Count} & \textbf{\%} & \textbf{Count} & \textbf{\%} \\
\midrule
Mostly Native (Telugu/Nepali) & 1,917 & 95.85 & 850 & 34.0 \\
Code-mixed & 80 & 4.00 & 820 & 32.8 \\
Universal / Other & 3 & 0.15 & 50 & 2.0 \\
\addlinespace
\textbf{Total} & \textbf{2,000} & \textbf{100.0} & \textbf{2,500} & \textbf{100.0} \\
\bottomrule
\end{tabular}
\end{table}

\section{Experiments on the Dataset}\label{sec:experiments}

\begin{figure}[H]
    \centering
    \includegraphics[height=0.6\textheight]{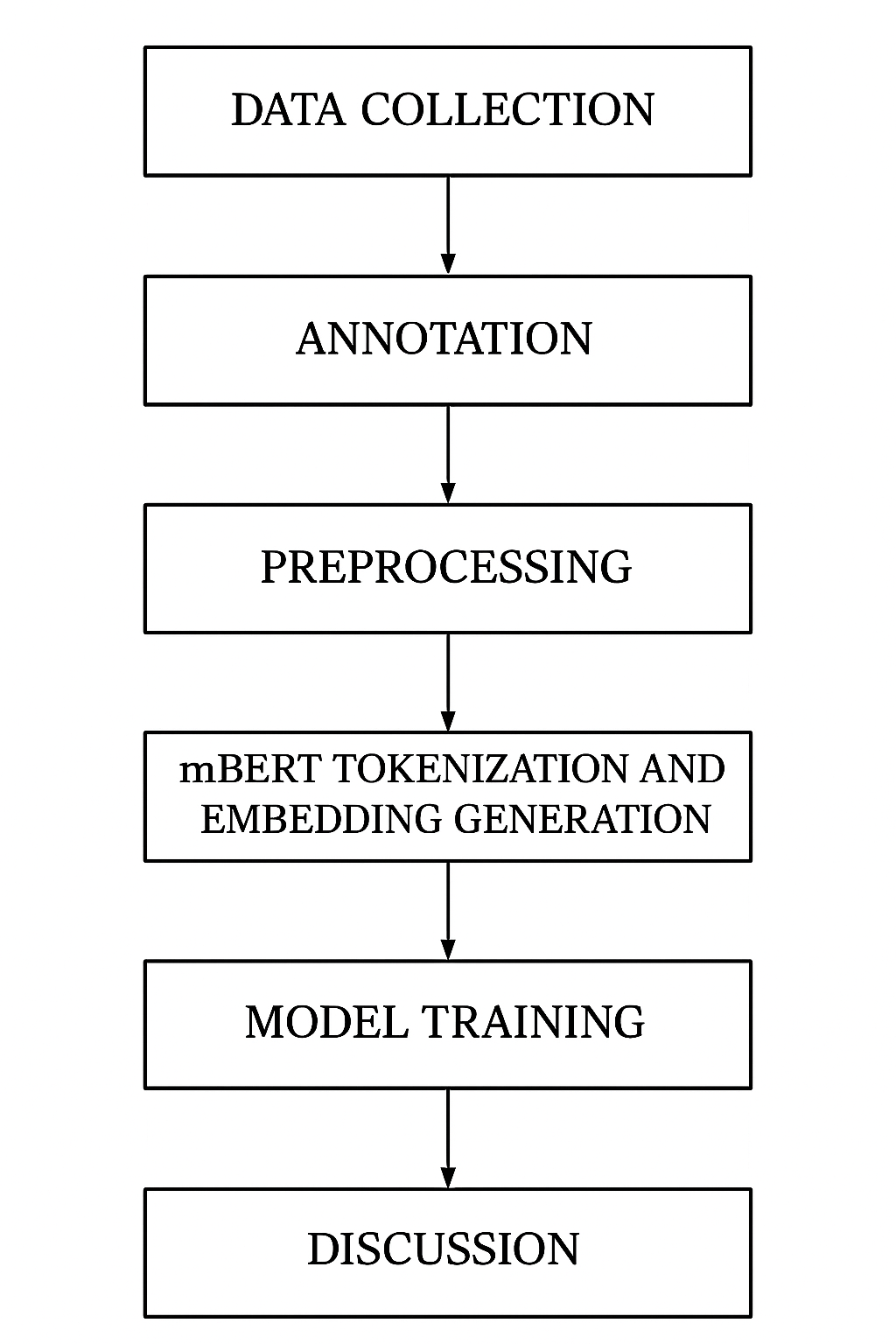}
    \caption{Workflow of Experimental setup}
    \label{fig:workflow}
\end{figure}

Figure ~\ref{fig:workflow} describes the complete workflow used for abusive language classification in code-mixed Nepali and Telugu corpora. The process starts with Data Collection, wherein the user-created content is harvested from social media websites. It is then followed by Annotation, where human annotators tag each sample as abusive or non-abusive to form a supervised learning corpus. Then comes the Preprocessing phase where the text is cleaned from noise like special characters, emojis, and duplicate symbols and linguistic patterns are normalized. Following this, the data passes through mBERT Tokenization and Embedding Generation, which employs multilingual BERT to tokenize the input and produce contextual embeddings specifically for every sentence. These embeddings are input into the Model Training step, wherein several machine learning and deep learning models are trained on the preprocessed data. The trained models are then evaluated using default classification metrics like Precision, Recall, F1-Score, and Accuracy. Lastly, Result Analysis and Model Comparison is done, involving statistical testing, to identify the best model for the task and to make meaningful conclusions from the experimental results.

\subsection{Preprocessing}
Due to the inherent linguistic differences and unique difficulties that each language posed, separate preprocessing pipelines were formulated to efficiently clean, normalize, and ready the text for downstream analysis and model training. This subsection explains the preprocessing steps adopted for the gathered Telugu-English and Nepali-English code-mixed data.
\subsubsection{Telugu Text Preprocessing}

The Telugu text preprocessing starts with Unicode normalization to deal with any encoding inconsistency. The text is normalized in the NFKC form so that characters are represented consistently. The text is then cleaned after normalization such that non-Telugu characters, special symbols, emojis, and unnecessary punctuation are eliminated from the text using regular expressions. Only Telugu characters in the Unicode range (\texttt{\textbackslash u0C00--\textbackslash u0C7F}) are preserved, and additional spaces, tabs, and newline characters are removed.

While Telugu is not case-sensitive, any Latin-script text (e.g., foreign words, hashtags) is changed to lowercase for consistency. The text is then tokenized using the mBERT tokenizer, which breaks the text into subword units and produces 768-dimensional embeddings for each token. This produces high-dimensional feature vectors that are passed into downstream classification models.

Stopwords are also filtered out of the text so that frequently occurring words with minimal semantic meaning are not able to influence the classification. The last preprocessed text is then available for feature extraction and training of models.

\subsubsection{Nepali Text Preprocessing}

The preprocessing of Nepali text is similar to that of Telugu, beginning with Unicode normalization using the NFKC form in order to ensure uniform character encoding. Non-Nepali characters, unwanted punctuation, emoticons, and non-task-relevant numbers are removed from the text. Only Nepali characters within the Unicode range (\texttt{\textbackslash u0900--\textbackslash u097F}) are retained, and additional spaces, tabs, and newline characters are eliminated using regular expressions.

Since there is no casing involved in Nepali text, Latin-script words such as foreign names or hashtags are converted to lowercase to maintain consistency. The preprocessed Nepali text is then tokenized using the mBERT tokenizer, which segments the text into subword pieces and generates 768-dimensional embeddings for each token. These embeddings are subsequently used as input features by the classification models.

As part of preprocessing, stopword removal is also performed to eliminate frequently occurring words such as ``the,'' ``is,'' or ``in,'' which do not carry significant semantic content for the classification task. The preprocessed text is then ready for analysis and model training.

\subsection{Model Training}
Here, we present the machine learning models selected for the task of classification, giving an idea of the training methods and strategies employed to identify abusive language successfully.
\subsubsection{Logistic Regression}
Logistic Regression is a straightforward but powerful binary classification model. It is most useful when there is a linear relationship between features and the target variable, and therefore can be a suitable option for text classification problems when the relationships are not very complex. The model is trained using 768-dimensional mBERT embeddings, which are numerical vectors that capture the semantic meaning of the text. These embeddings are then fed into the model so that it can predict whether the text is abusive or not based on the coefficients it learns during training \cite{devlin2019bert} \cite {ruder2019survey}.

\subsubsection{SVM}
SVM is employed for its capacity to deal with high-dimensional spaces well, and it is suitable for text classification where the input data (embeddings) is of high dimension. SVM operates by determining the best hyperplane that can divide the two classes. The 768-dimensional mBERT embeddings are input as features to the model, and SVM utilizes these to calculate the best boundary that separates abusive from non-abusive text \cite{joachims1998text}.

\subsubsection{Random Forest}
Random Forest is an ensemble learning technique that functions by creating several decision trees, enhancing classification accuracy and preventing overfitting \cite{breiman2001random}. It is particularly effective when there are intricate, non-linear patterns in the data. The model takes 768-dimensional mBERT embeddings as input, where the embedding of each text is used as a feature for the decision trees. The trees’ ensemble votes as a whole to determine whether the text is abusive or not.

\subsubsection{Neural Network}
The Neural Network (MLP) is employed because it can represent complex, non-linear relationships within the data. This is particularly beneficial in text classification, where word and context relationships are not necessarily linear. The 768-dimensional mBERT embeddings are fed as input features to the network. The network feeds these embeddings through several layers, learning complex patterns to classify the text as abusive or non-abusive.

\subsubsection{CNN}
CNNs are good at pulling out local patterns from data and are, therefore, well-suited for tasks where local dependencies (such as word sequences) are critical, e.g., text classification. The mBERT 768-dimensional embeddings are input to the CNN model, with the convolutional layers performing filtering to pick up on salient local patterns (e.g., particular word combinations or n-grams) and then labeling the text as abusive or not abusive based on these features.
\subsubsection{LSTM}
LSTM is used for sequential data, where the relationships between elements depend on the sequence and context. It is particularly effective in capturing long-range dependencies in text, making it ideal for understanding context in sentences. The 768-dimensional mBERT embeddings are fed sequentially into the LSTM, which processes the embeddings and captures temporal relationships between words to classify the text as abusive or non-abusive.

\subsubsection{LLM}
The methodology entails a systematic investigation of hyperparameter setups to assess the impact of learning rate and training time on the performance of transformer-based classifiers. Several fine-tuning experiments were performed with different learning rates (1e-5, 3e-5, and 5e-5) for different numbers of training epochs. Model performance was measured in terms of classification accuracy in order to determine optimal setups and to investigate training stability \cite{dodge2020fine}. This perspective facilitates a better understanding of how fine-tuning dynamics affect performance, particularly in low-resource and cross-lingual scenarios, and points to the necessity of selecting hyperparameters with caution—specifically learning rate adjustment—to achieve strong model performance.

\section{Results}\label{sec:results}
This section presents a comparative discussion of the model performance on both Nepali and Telugu datasets, analyzing differences in results and how linguistic variations affect classification accuracy.
\subsection{Nepali Dataset}
This sub-section provides the results of the evaluation of the machine learning models used on the Nepali dataset, showing their performance in identifying abusive language in Nepali-English code-mixed text.

\subsubsection{Result on Logistic Regression}

\begin{figure}[H]
    \centering
    \includegraphics[width=\linewidth]{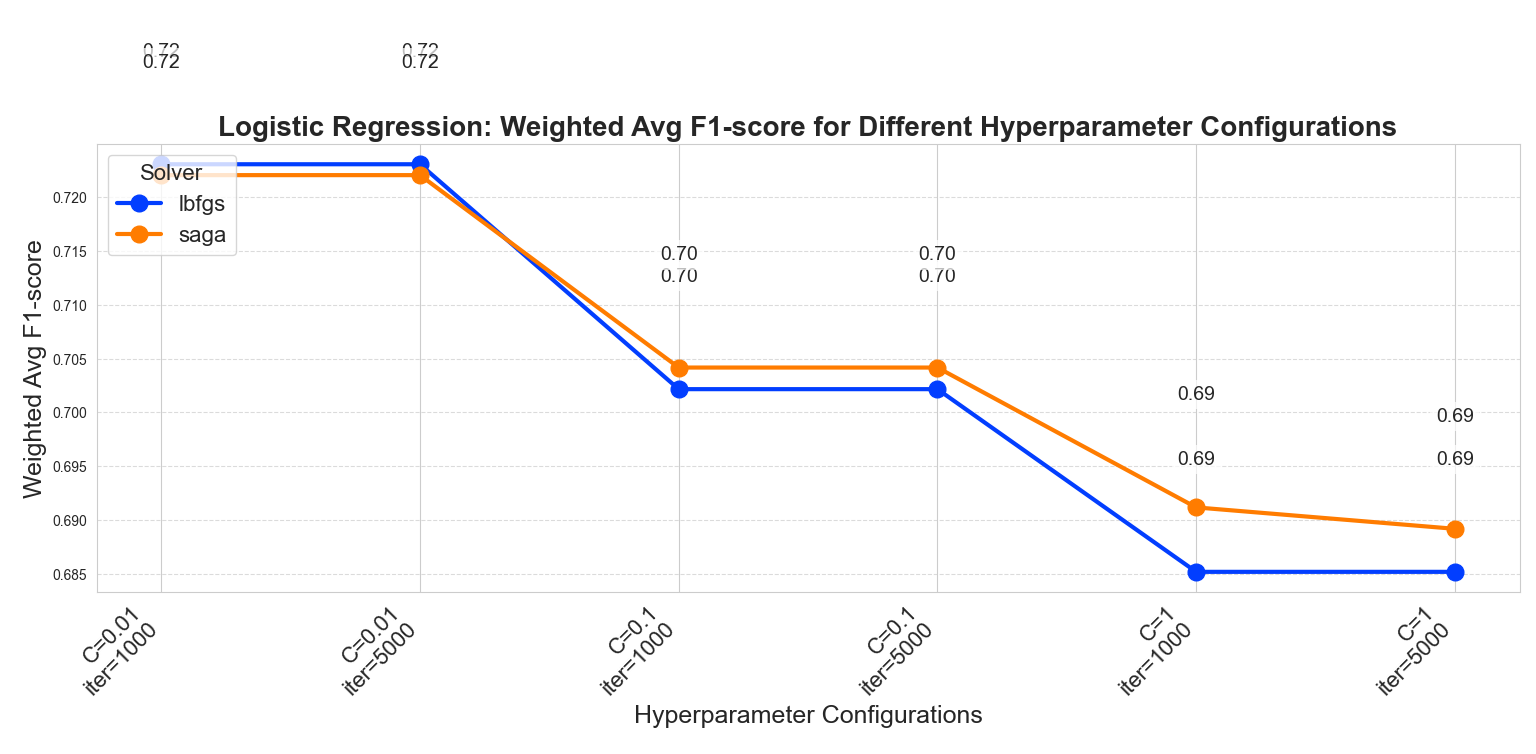}
    \caption{ F1-Score Trends for Logistic Regression}
    \label{fig:fullwidth}
\end{figure}

The \textbf{Logistic Regression model with lbfgs solver} produced the highest \textbf{weighted F1-score of 0.723} (accuracy: 72.3\%) with the following parameters: 
\texttt{C=0.01}, \texttt{penalty='l2'}, class\_weight='balanced', and \texttt{max\_iter=1000}. 
Lower regularization strength (\texttt{C=0.01}) performed consistently better than higher values (\texttt{C=0.1}, \texttt{C=1}), with F1-scores falling to 0.702 and 0.685, respectively, 
showing weaker regularization better handles bias-variance trade-offs for this dataset. 
The \textbf{lbfgs solver} beat saga by a little (F1=0.723 vs. 0.722 at \texttt{C=0.01}), the probable reason for its efficiency at L2 penalty. 
Class weighting (balanced) guaranteed unbiased performance, with the minority class (1) F1-scores being no more than 2\% lower than the majority class (0). 
Doubling \texttt{max\_iter} to 5000 produced identical results to the 1000 iterations, vindicating the efficiency of convergence. 
These results point out the importance of moderate regularization (\texttt{C=0.01}) and solver choice (\texttt{lbfgs}) in the optimization of logistic regression in well-balanced classification problems.
.

\subsubsection{Result on SVM Model}

\begin{figure}[H]
    \centering
    \includegraphics[width=\linewidth]{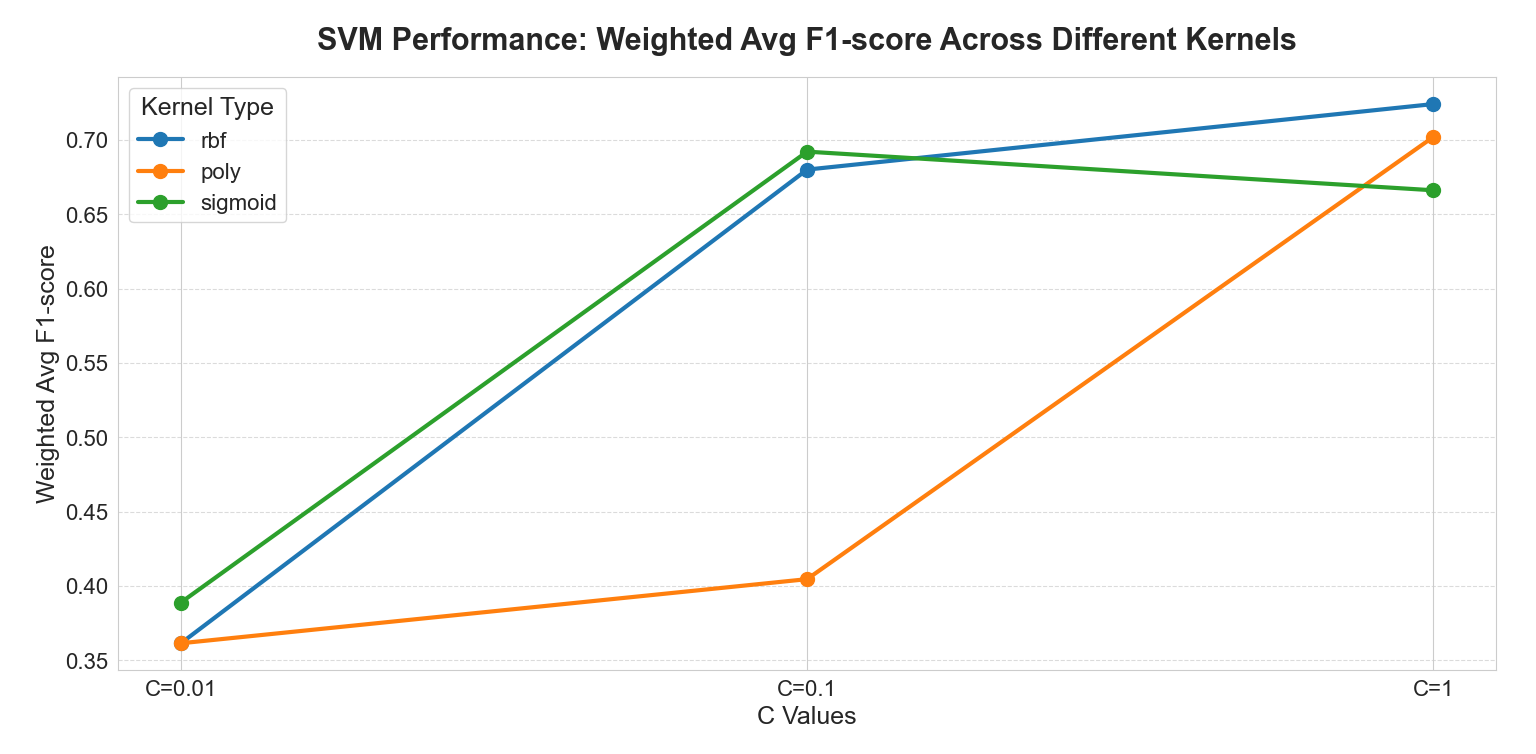}
    \caption{F1-Score Trends for SVM}
    \label{fig:fullwidth}
\end{figure}

The \textbf{SVM model with RBF kernel} and \textbf{C=1} had the best weighted F1-score of \textbf{0.724} (accuracy: 72.6\%), showing the radial basis function's ability to capture non-linear decision boundaries when combined with moderate regularization. 
Lower C values (e.g., 0.01) for all kernels resulted in extreme underfitting, especially for class 1, with F1-scores close to zero, indicating poor generalization. 
Increasing C enhanced performance, but very high values (\texttt{C=1} for sigmoid kernel) compromised results, indicating kernel-dependent sensitivity to regularization. 
The \textbf{poly kernel} (\texttt{C=1}) was second-best (F1=0.702), with higher precision for class 1 (78.5\%), and the \textbf{sigmoid kernel} reached a peak at \texttt{C=0.1} (F1=0.692), reflecting limited applicability to the dataset. 
Of particular interest, low C values showed class 0 recall bias (up to 99.6\%), well in excess of class 1, pointing to class imbalance issues, which were reduced by increasing C values in RBF and poly setups. 
These findings strongly support the need for the selection of kernels and regularization parameter tuning in SVMs, with the RBF kernel (\texttt{C=1}, \texttt{gamma='scale'}) being the best setup for balanced performance on this task.

\subsubsection{Result on Random Forest}
\begin{figure}[H]
    \centering
    \includegraphics[width=\linewidth]{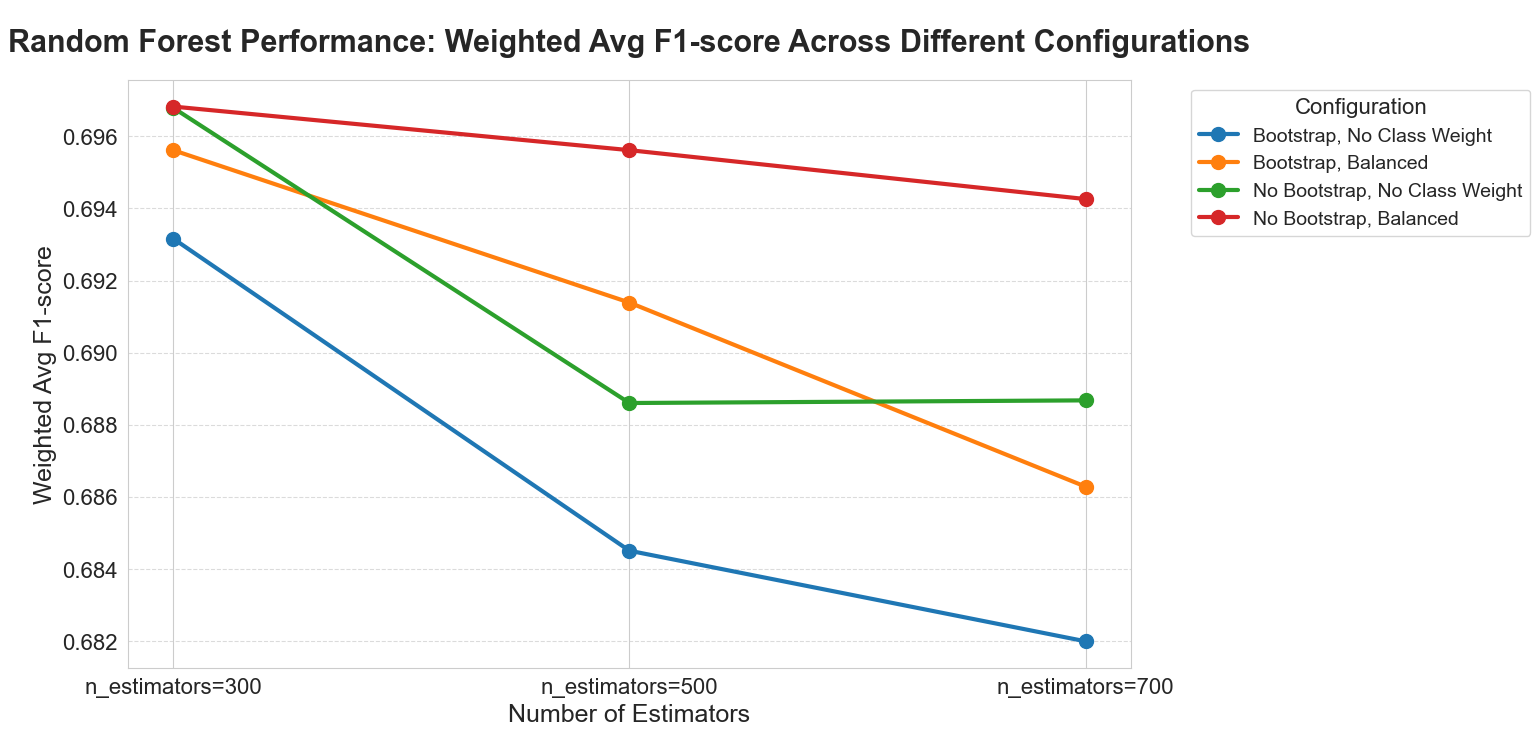}
    \caption{F1-Score Trends for Random Forest}
    \label{fig:fullwidth}
\end{figure}

The Random Forest model with 300 trees and disabled bootstrap sampling achieved the highest weighted F1-score of 0.697 (accuracy: 70.1\%), using the following parameters: 
\texttt{n\_estimators=300, bootstrap=False, max\_depth=20,} \\
\texttt{max\_features='sqrt', min\_samples\_leaf=2, min\_samples\_split=5,} \texttt{class\_weight=None}.
Disabling bootstrap sampling slightly improved performance when compared to bootstrapped configurations (e.g., 300 trees with no bootstrap vs. bootstrap: F1=0.697 vs. 0.693), 
likely due to reduced overfitting by utilizing the full dataset for tree construction. Increasing the number of trees beyond 300 (e.g., 500 or 700) resulted in a degradation of performance (F1=0.689--0.697), 
suggesting diminishing returns with larger ensembles. Class balancing (\texttt{class\_weight='balanced'}) marginally improved the recall for the minority class (class 1), 
but at the cost of reduced overall precision, indicating minimal impact from class imbalance. Fixed hyperparameters such as \texttt{max\_depth=20} and lower values for \texttt{min\_samples\_leaf} 
and \texttt{min\_samples\_split} (2 and 5, respectively) provided sufficient model complexity without overfitting, as performance remained stable across configurations. 
These results highlight the importance of optimizing bootstrap sampling and the number of trees, with smaller ensembles (300 trees) and full dataset utilization offering the best trade-off between bias and variance for this task.

\subsubsection{Result on Neural Network}

\begin{figure}[H]
    \centering
    \includegraphics[width=\linewidth]{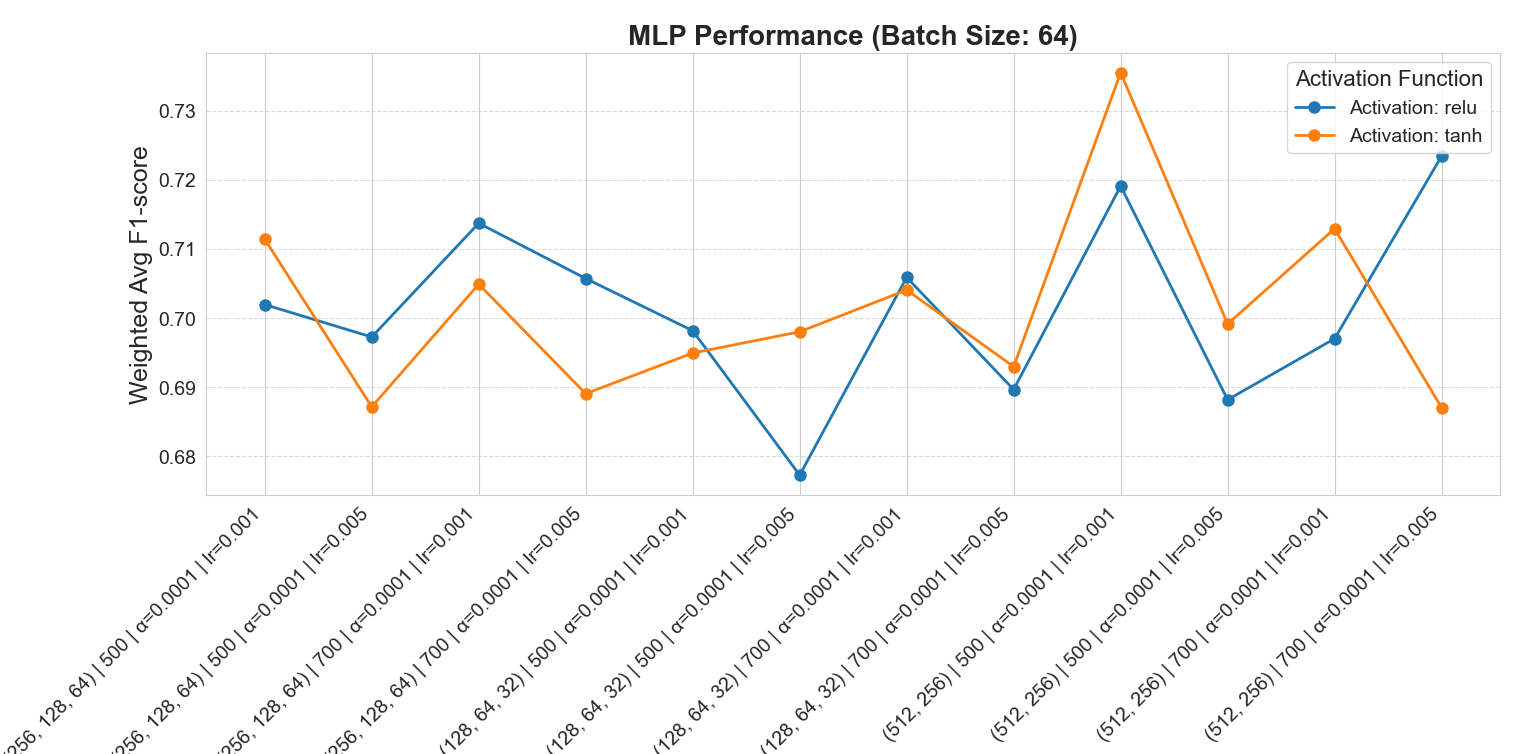}
    \caption{F1-Score Trends for Neural Network with Batch size: 64}
    \label{fig:fullwidth}
\end{figure}

\begin{figure}[h]
    \centering
    \includegraphics[width=\linewidth]{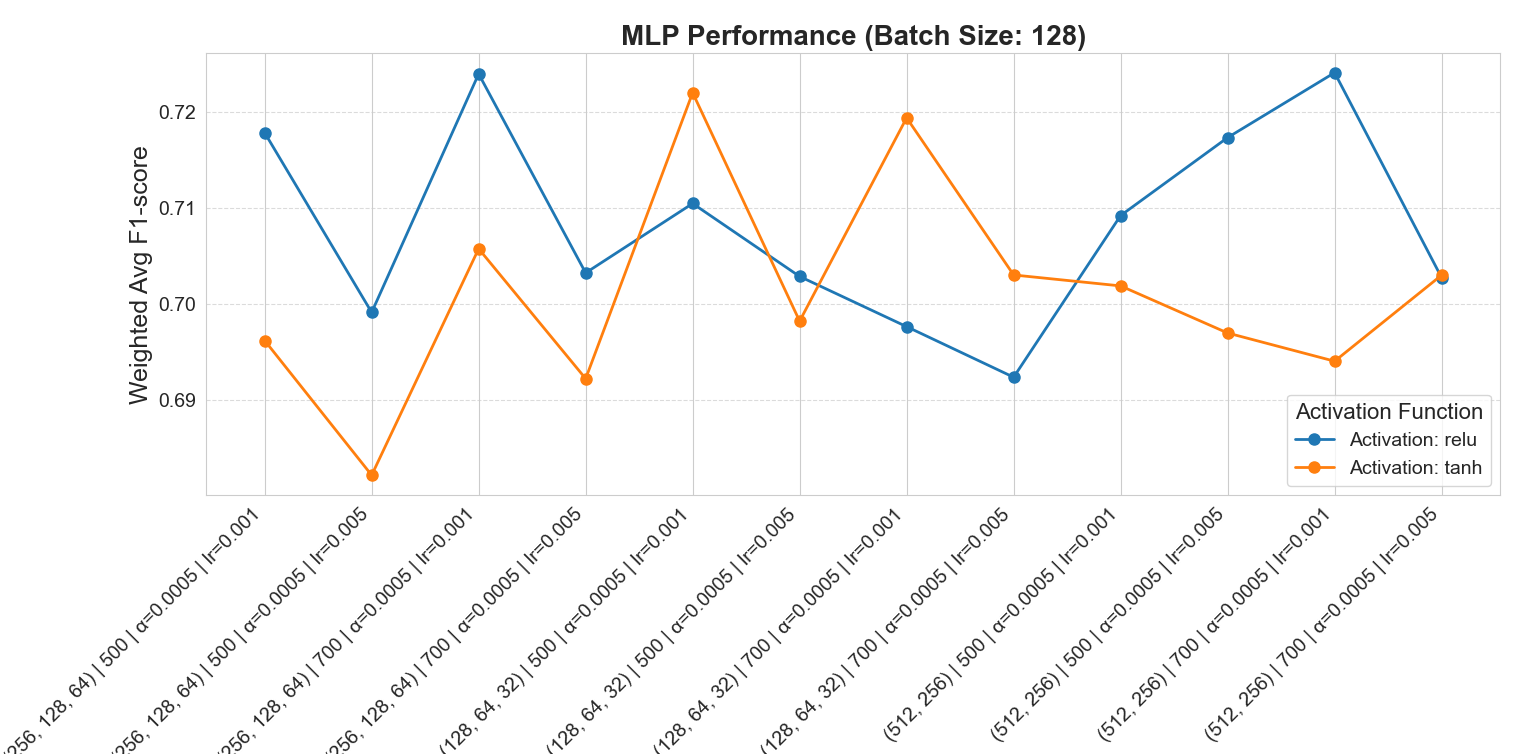}
    \caption{F1-Score Trends for Neural Network with Batch size: 128}
    \label{fig:second_image}
\end{figure}

The Neural Network tanh activation function and Adam optimizer achieved the highest weighted F1-score of 0.736 (accuracy: 73.6\%), having the following parameters: 
\texttt{hidden\_layer\_sizes=(512, 256), max\_iter=500, batch\_size=64, alpha=0.0001, learning\_rate\_init=0.001}.
Deeper architectures having 512-256 layers outperformed shallower networks with 128-64-32 layers, suggesting enhanced hierarchical feature extraction. 
The tanh activation function consistently surpassed ReLU in larger models (e.g., 73.6\% vs. 71.9\% accuracy), likely due to its zero-centered output stabilizing gradients. 
Smaller batch sizes (i.e., \texttt{batch\_size}=64), paired with moderate learning rates (0.001), improved convergence, while higher rates (0.005) destabilized training, reducing F1-scores by 2--4\%. 
Lower L2 regularization (\texttt{alpha}=0.0001) outperformed stronger regularization (\texttt{alpha}=0.0005), indicating minimal overfitting despite model complexity. 
Extending training iterations beyond 500 (\texttt{max\_iter}=700) yielded marginal gains, as early stopping effectively prevented overfitting. 
These results underscore the synergy of architectural depth, activation choice (tanh), and adaptive optimization (Adam) in balancing feature learning and generalization for optimal performance.

\subsubsection{Result on CNN}

\begin{figure}[H]
    \centering
    \includegraphics[width=\linewidth]{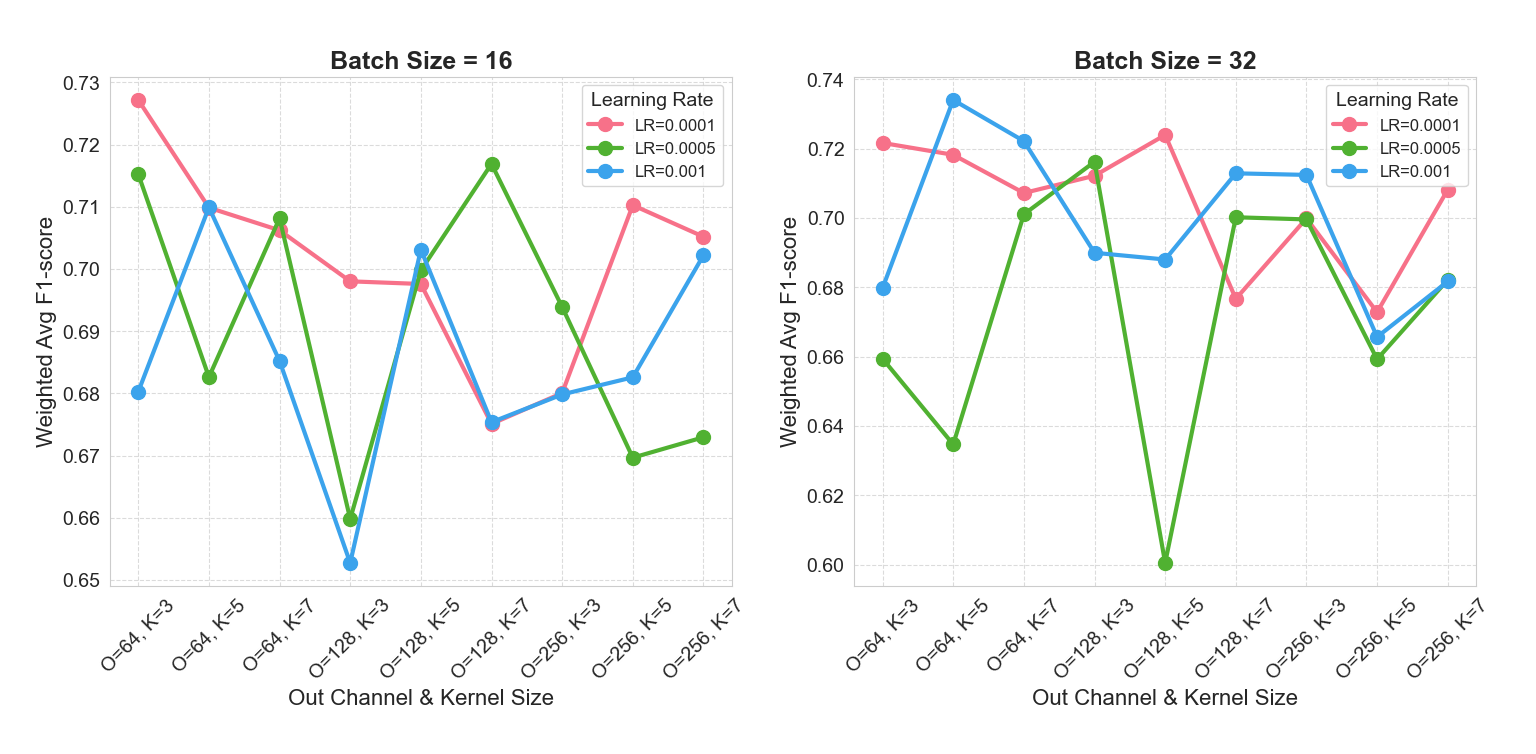}
    \caption{F1-Score Trends for CNN}
    \label{fig:fullwidth}
\end{figure}

The highest-scoring model has a \textbf{weighted F1-score of 0.7340} (accuracy: 73.5\%) using the following parameters: 
\texttt{out\_channels=64}, kernel\_size=5, dropout\_rate=0.3, learning\_rate=0.001, and \texttt{batch\_size=32}. 
This setting implies that smaller networks (64 channels) with an intermediate kernel size (5) and a more elevated learning rate (0.001) work well to equilibrate feature extraction and generalization. 
Interestingly, growing \texttt{out\_channels} to 128 or 256 tended to result in incremental or diminished performance, suggesting diminishing returns for greater models. 
\texttt{kernel\_size=3} or 7 models were inconsistent, but kernel size 5 consistently produced competitive results. 
Reduced learning rates (0.0001) combined with lower batch sizes (16) sometimes offered improved stability but at the expense of the F1 score, and increased learning rates (0.001) combined with batch size 32 had improved performance combined with optimal kernel sizes. 
Dropout (0.3) held steady, implying its usefulness in avoiding overfitting without learning detriment. 
Across the board, the relationship among kernel size, learning rate, and batch size had considerable effects on model performance, 
the best combination prioritizing moderate complexity and adaptive learning.

\subsubsection{Result on LSTM Model}

\begin{figure}[H]
    \centering
    \includegraphics[width=\linewidth]{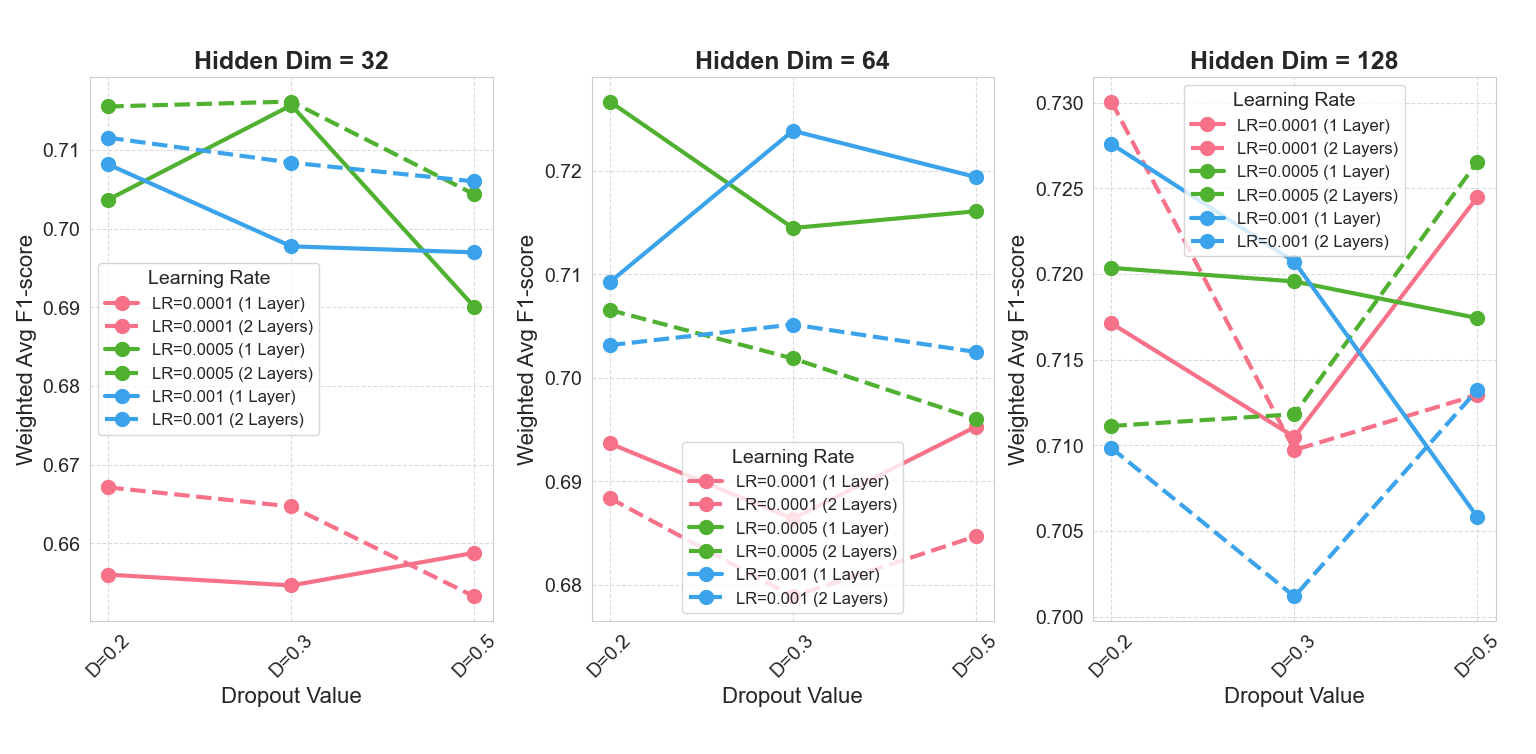}
    \caption{F1-Score Trends for LSTM}
    \label{fig:fullwidth}
\end{figure}

The LSTM model analysis indicates that architectural complexity and parameter tuning have a strong impact on performance, 
with the best weighted average F1-score (\textbf{0.7301}) and accuracy (\textbf{73.1\%}) being obtained by the model with 
\texttt{hidden\_dim=128}, \texttt{num\_layers=2}, \texttt{bidirectional=True}, \texttt{dropout=0.2}, learning\_rate=0.0001, and \texttt{batch\_size=32}. 
Bidirectional LSTMs with two layers performed better than more basic architectures throughout, demonstrating their capacity to capture contextual dependencies in sequential data. 
Smaller dropout rates (0.2–0.3) equilibrated regularization and model capacity, while higher rates (e.g., 0.5) compromised performance by as much as \textbf{2.3\% F1}, presumably from too much regularization. 
Lower learning rates (\texttt{learning\_rate=0.0001}) improved stability on deeper architectures, while higher rates (\texttt{0.001}) produced unstable convergence, especially in bidirectional models. 
Higher hidden dimensions (\texttt{hidden\_dim=128} compared to 32/64) boosted F1 by \textbf{4–6\%}, highlighting their ability to capture complex patterns. 
Although smaller batch sizes (\texttt{batch\_size=16}) slightly favored single-layer models, bidirectional architectures alleviated this disparity, sustaining strong performance with batch sizes of 32. 
The poorest performers (F1 less than 0.66), including models with \texttt{hidden\_dim=32} and high dropout (0.5), were plagued by low capacity and over-regularization. 
These results highlight the need to balance architectural depth, regularization, and learning dynamics to maximize LSTM performance in sequential classification tasks.

\subsubsection{Result on LLM}
Analysis showed critical dependencies among learning rates, training length, and model performance, 
with the best configuration (\texttt{1e-5 learning rate}, 4 epochs) achieving 78.3\% accuracy, highlighting the effectiveness of smaller learning rates for stable fine-tuning. 
Models with higher rates showed extreme degradation: \texttt{3e-5} demonstrated volatility (peak 71.2\% at 3 epochs, decreasing to 65.2\% by epoch 4), 
whereas \texttt{5e-5} catastrophically failed (53.9\% for all epochs), showing optimization instability. 
Interestingly, longer training benefited only the \texttt{1e-5} setting (73.6\% $\to$ 78.3\% after 2–4 epochs), 
while higher rates plateaued or declined, indicating overfitting or unstable gradient updates. 
These findings highlight the importance of careful hyperparameter tuning, especially learning rate adjustment, for transformer-based classifiers, 
with lower rates (\texttt{1e-5}) being most resilient to extended training. 
The dramatic 24.4\% accuracy gap between optimal (78.3\%) and worst (53.9\%) settings sheds light on the biased effect of parameter tuning versus architectural design in low-resource settings, 
calling for principled validation of learning dynamics under cross-lingual transfer tasks.

\subsection{Telegu Dataset}
in this sub-section, we present the performance results of the machine learning models over the Telugu dataset, with emphasis on their capacity to identify abusive language in Telugu-English code-mixed text.

\subsubsection{Logistic Regression}

\begin{figure}[H]
    \centering
    \includegraphics[width=\linewidth]{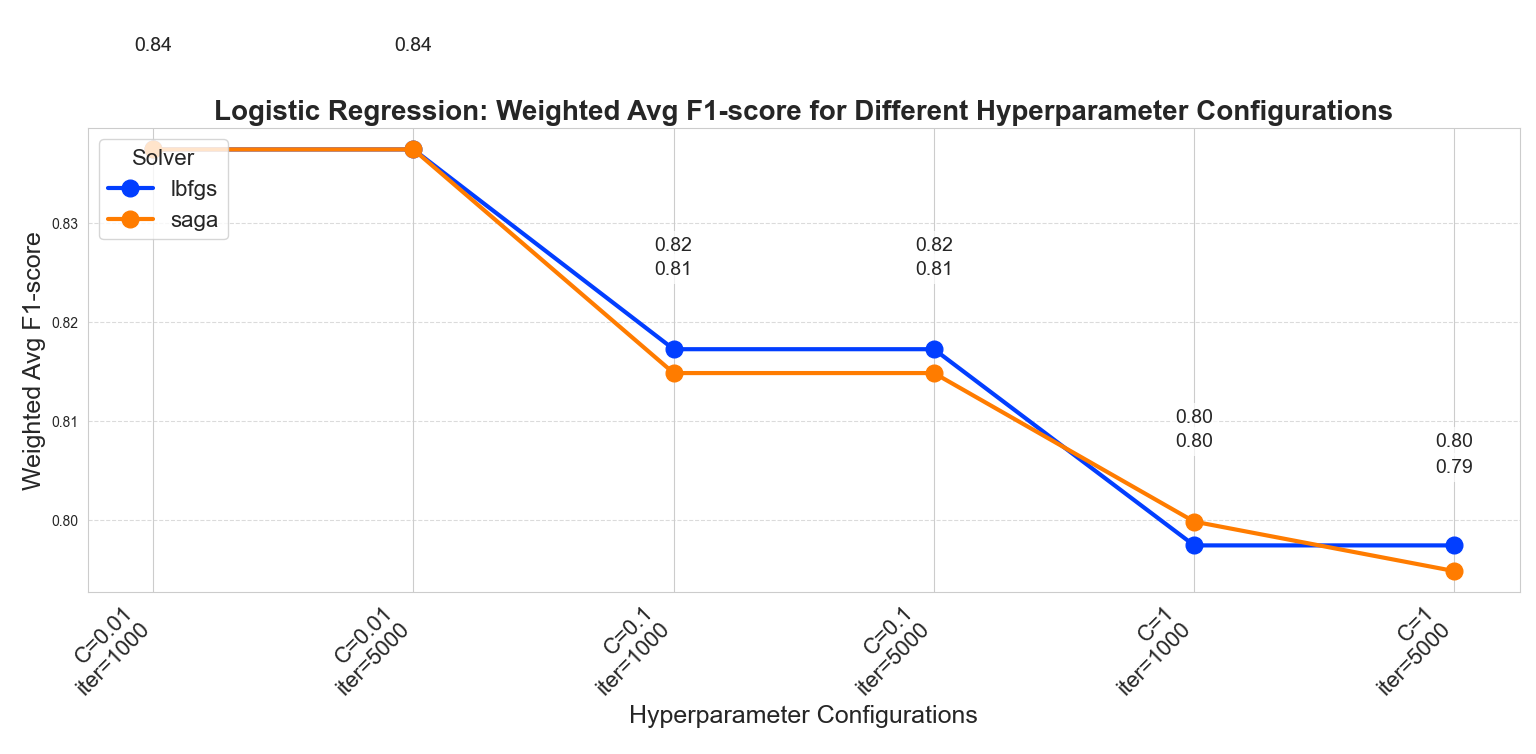}
    \caption{F1-Score Trends for Logistic Regression}
    \label{fig:fullwidth}
\end{figure}

The models show that \textbf{lower regularization strength (C=0.01)} produces the best performance, with a \textbf{weighted F1-score of 0.8374} and \textbf{83.75\% accuracy} across all solvers (\texttt{lbfgs}, \texttt{saga}) and iterations (\texttt{max\_iter=1000} or \texttt{5000}). 
This implies good generalizability with balanced class weights, which will probably counteract overfitting even with the weaker penalty. 
On the other hand, increased regularization (\texttt{C=1}) decreases performance (F1: \textbf{0.7948}, accuracy: \textbf{79.5\%}), implying underfitting. 
Of note, \textbf{solver choice} and \textbf{max\_iter} had minimal effect for \texttt{C=0.01} since all settings with these parameters converged to the same outcomes, suggesting optimization stability. 
The optimal model is obtained with \texttt{C=0.01}, \texttt{penalty='l2'}, \texttt{class\_weight='balanced'}, \texttt{solver='lbfgs'} (or \texttt{saga}), and \texttt{max\_iter=1000}, 
which produces stable precision (0.8608) and recall (0.8644) for the majority class while providing balanced performance for the minority class (F1: 0.8012). 
This demonstrates the efficacy of moderate regularization and class balancing for imbalanced datasets.

\subsubsection{Result on SVM}

\begin{figure}[H]
    \centering
    \includegraphics[width=\linewidth]{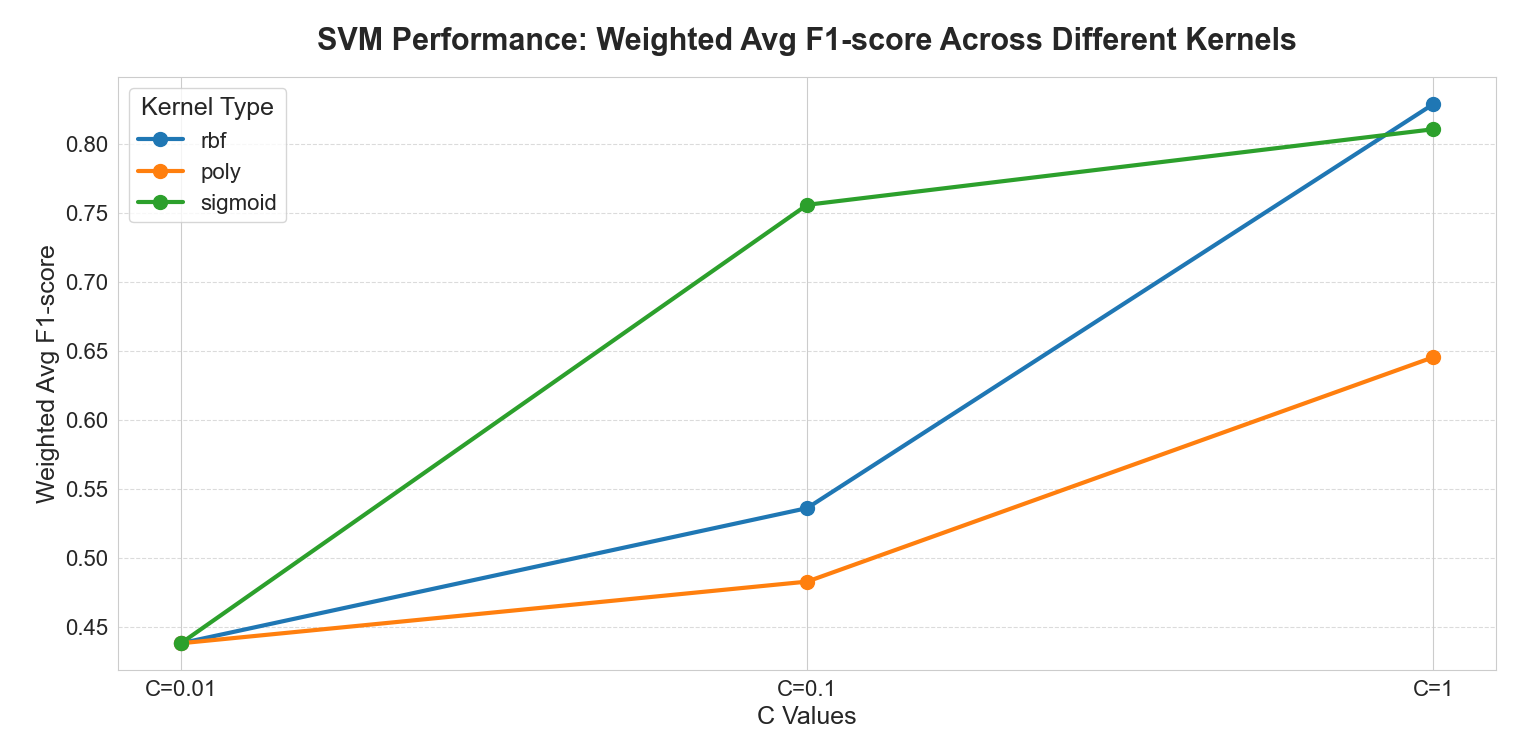}
    \caption{F1-Score Trends for SVM}
    \label{fig:fullwidth}
\end{figure}

Experiments compared SVM models with different kernels and strengths of regularization (\texttt{C}) for binary classification. 
Best-performing \textbf{SVM (RBF Kernel, C=1)} returned an \textbf{83.25\% accuracy} and a \textbf{82.86\% weighted F1-score} under parameters such as 
\texttt{kernel='rbf'}, \texttt{C=1}, and \texttt{gamma='scale'}. 
Evidence showed that RBF kernel tended to outperform polynomial (\texttt{poly}) and sigmoid kernels across instances, presumably by capturing complex non-linear decision surfaces without overfitting. 
Stronger regularization strength (\texttt{C=1}) enhanced performance in all kernels, striking the margin flexibility vs. generalization balance, 
while lower strengths (\texttt{C=0.01} or \texttt{0.1}) resulted in underfitting, as observed in models that predicted only the most frequent class (e.g., 
\texttt{poly} and \texttt{sigmoid} with \texttt{C=0.01}: F1 $\le$ 43.79\%). 
The sigmoid kernel performed reasonably well (\texttt{C=1}: F1=81.04\%), but was challenged by class imbalance relative to RBF. 
Of particular note, the polynomial kernel had low precision for the minority class (\texttt{poly, C=1}: precision=90.38\% for class 0 but recall=28.66\%), 
which means overfitting to certain features. 
The success of the best model highlights the value of kernel choice and regularization: the RBF kernel's flexibility with respect to data geometry, 
along with \texttt{C=1}, best traded off margin violations against discriminative power. 
These results emphasize that SVMs with RBF kernels and balanced regularization are stable for imbalanced classification problems when tuned to avoid overfitting.

\subsubsection{Result on Random Forest}

\begin{figure}[H]
    \centering
    \includegraphics[width=\linewidth]{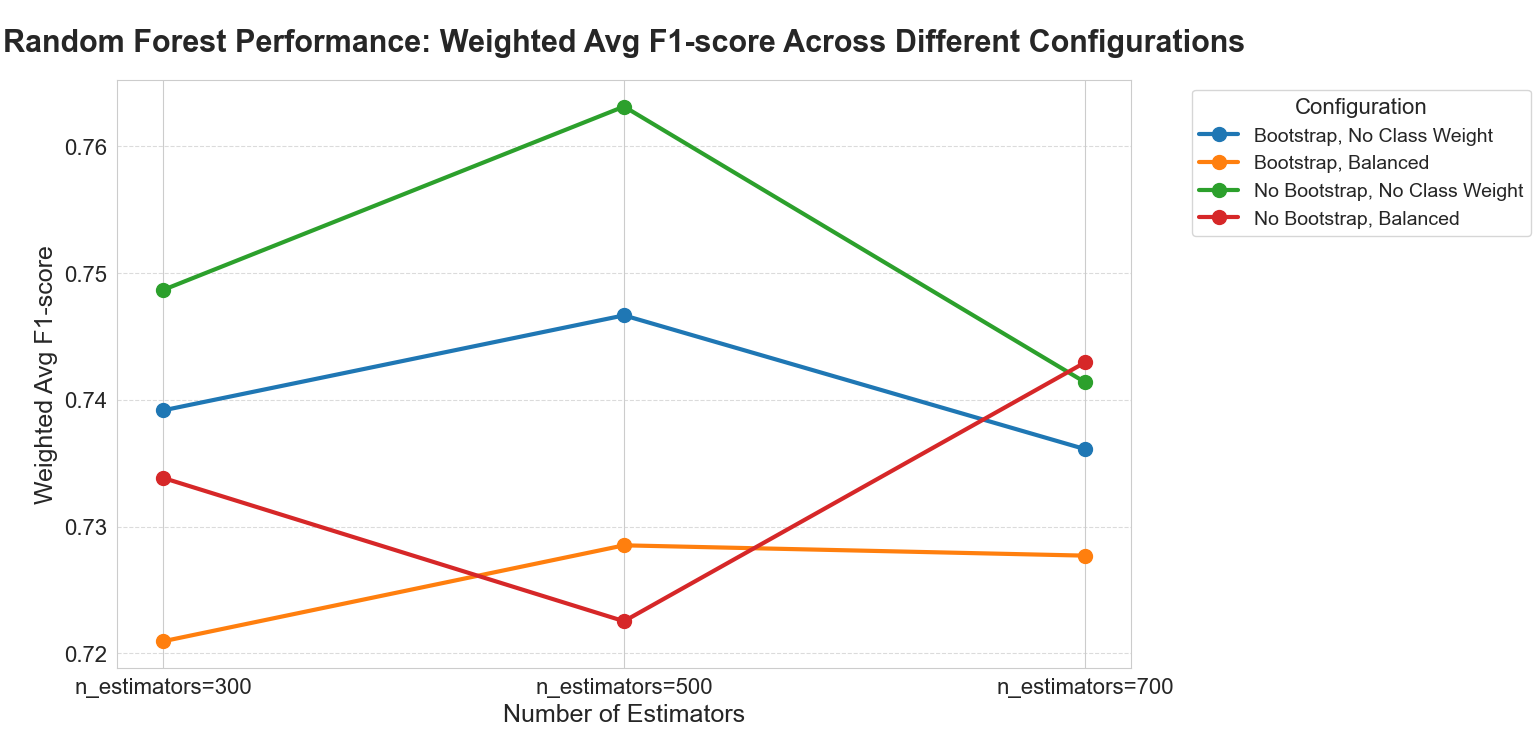}
    \caption{F1-Score Trends for Random Forest}
    \label{fig:fullwidth}
\end{figure}

The experiments compared Random Forest models with different ensemble sizes (\texttt{n\_estimators}), bootstrapping approaches, and class-balancing methods for a binary classification problem. 
The best-performing model, \textbf{Random Forest (500 Trees, No Bootstrap)}, had an \textbf{accuracy of 77.50\%} and a \textbf{weighted F1-score 76.31\%} using parameter such as n\_estimators=500, bootstrap=False, max\_depth=20, max\_features='sqrt', min\_samples\_leaf=2,
\\ min\_samples\_split=5, and \texttt{class\_weight=None}. Analysis showed that bootstrapping was disabled (\texttt{bootstrap=False}) consistently to always enhance performance in all ensemble sizes, presumably from less bias caused by using the entire dataset in each tree. 
Larger ensembles (\texttt{n\_estimators=500}) slightly performed better than smaller ones (\texttt{n\_estimators=300/700}), indicating a balance between variance reduction and computational complexity. 
Class weighting (\texttt{class\_weight='balanced'}) compromised F1-scores (e.g., \texttt{500 Trees, \\ Bootstrap\_Balanced}: F1=72.85\% vs. \texttt{500 Trees, No Bootstrap}: F1=76.31\%), 
demonstrating the use of default class weighting adequately addressed imbalance without penalizing the majority class too heavily. 
The \texttt{max\_depth=20} and \texttt{max\_features='sqrt'} settings offered suitable complexity and feature selection diversity, but using deeper trees could result in underfitting. 
Pertinently, models without bootstrapping had better recall for the minority class (e.g., \texttt{500 Trees, No Bootstrap}: recall=54.88\% for class 0 vs. \texttt{500 Trees, Bootstrap}: recall=52.44\%), 
highlighting their ability to effectively identify rare patterns. 
These results highlight the significance of ensemble size, bootstrapping approach, and conservative class weighting in optimizing Random Forests for imbalanced classification problems.

\subsubsection{Result on Neural Network}

\begin{figure}[H]
    \centering
    \includegraphics[width=\linewidth]{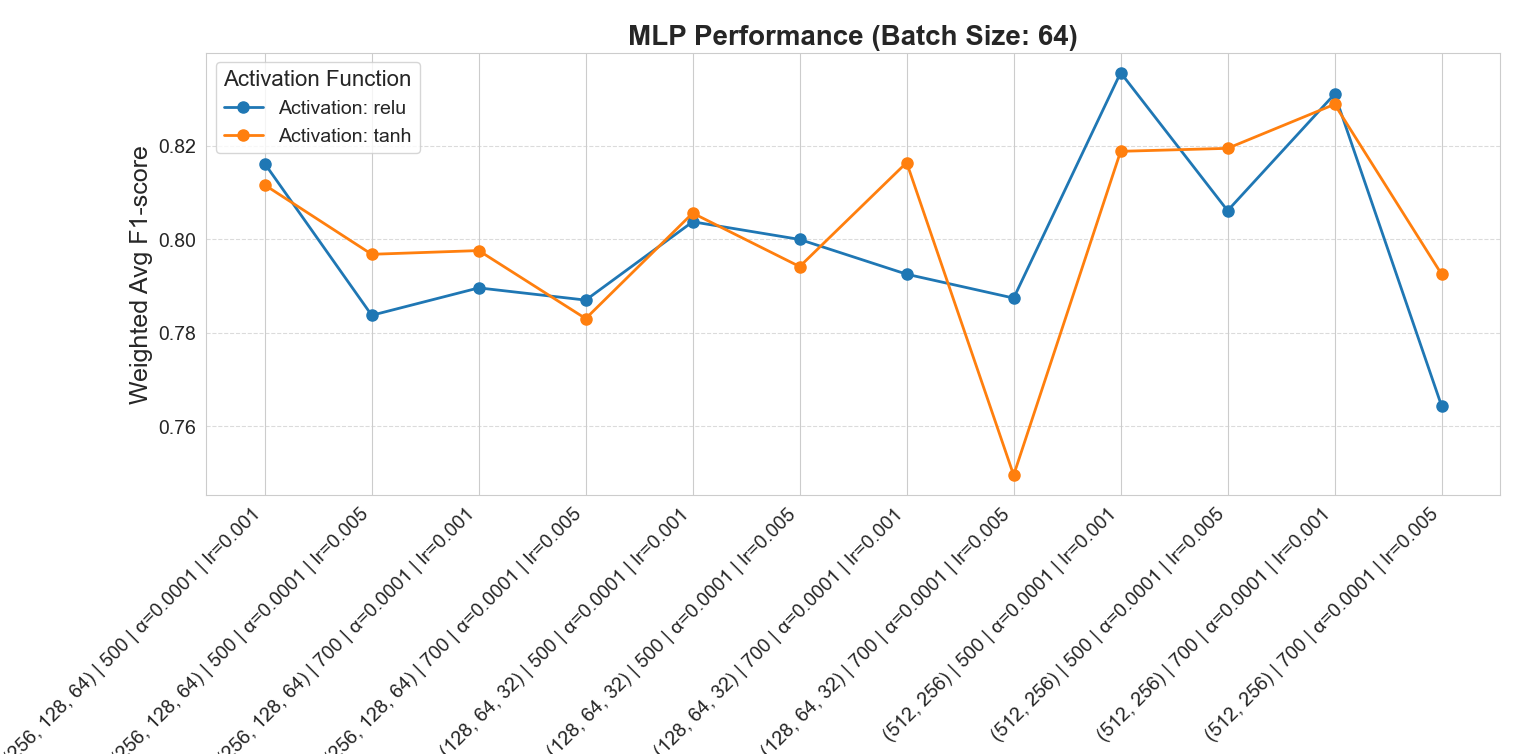}
    \caption{F1-Score Trends for Neural Network with Batch size: 64}
    \label{fig:fullwidth}
\end{figure}

\begin{figure}[h]
    \centering
    \includegraphics[width=\linewidth]{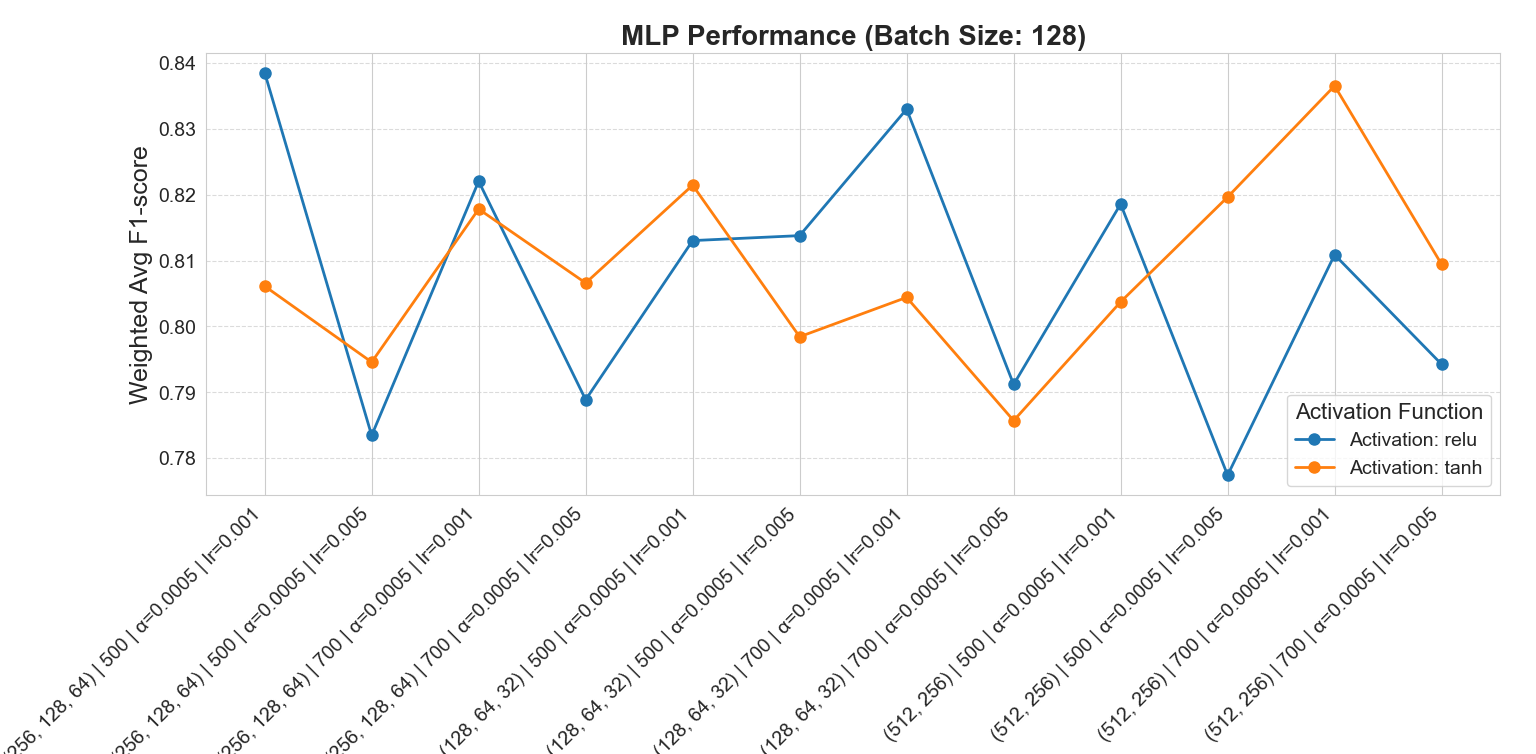}
    \caption{F1-Score Trends for Neural Network with Batch size: 128}
    \label{fig:second_image}
\end{figure}

The comparison of neural network models by weighted average F1-score provides key insights into parameter optimization. 
Models with deeper architectures (e.g., \textbf{(256, 128, 64)} hidden layers) tended to perform better than shallower ones, 
with greater stability and generalization. The \textbf{ReLU activation} function consistently provided better performance (weighted F1: up to \textbf{0.8795}) 
than \textbf{tanh}, presumably because it avoids vanishing gradients. 
Smaller learning rates (\texttt{0.001} vs. \texttt{0.005}) and adaptive learning rates enhanced convergence, 
whereas smaller batch sizes (\texttt{64} vs. \texttt{128}) sometimes minimized overfitting but at the expense of training stability. 
The highest performing model had a \textbf{weighted F1-score of 0.8795} and \textbf{accuracy of 84.25\%} with the following parameters: 
\texttt{hidden\_layer\_sizes=(256, 128, 64)}, \texttt{max\_iter=500}, \texttt{batch\_size=128}, \texttt{alpha=0.0005}, 
\texttt{learning\_rate\_init=0.001}, \texttt{activation=relu}, \texttt{solver=adam}. 
Interestingly, raising \texttt{max\_iter} above 500 did not result in meaningful improvements, indicating early convergence. 
These results emphasize the interaction between regularization (through \texttt{alpha=0.0005}) and architectural depth for maximizing discriminative ability in imbalanced classification problems.

\subsubsection{Result on CNN}

\begin{figure}[H]
    \centering
    \includegraphics[width=\linewidth]{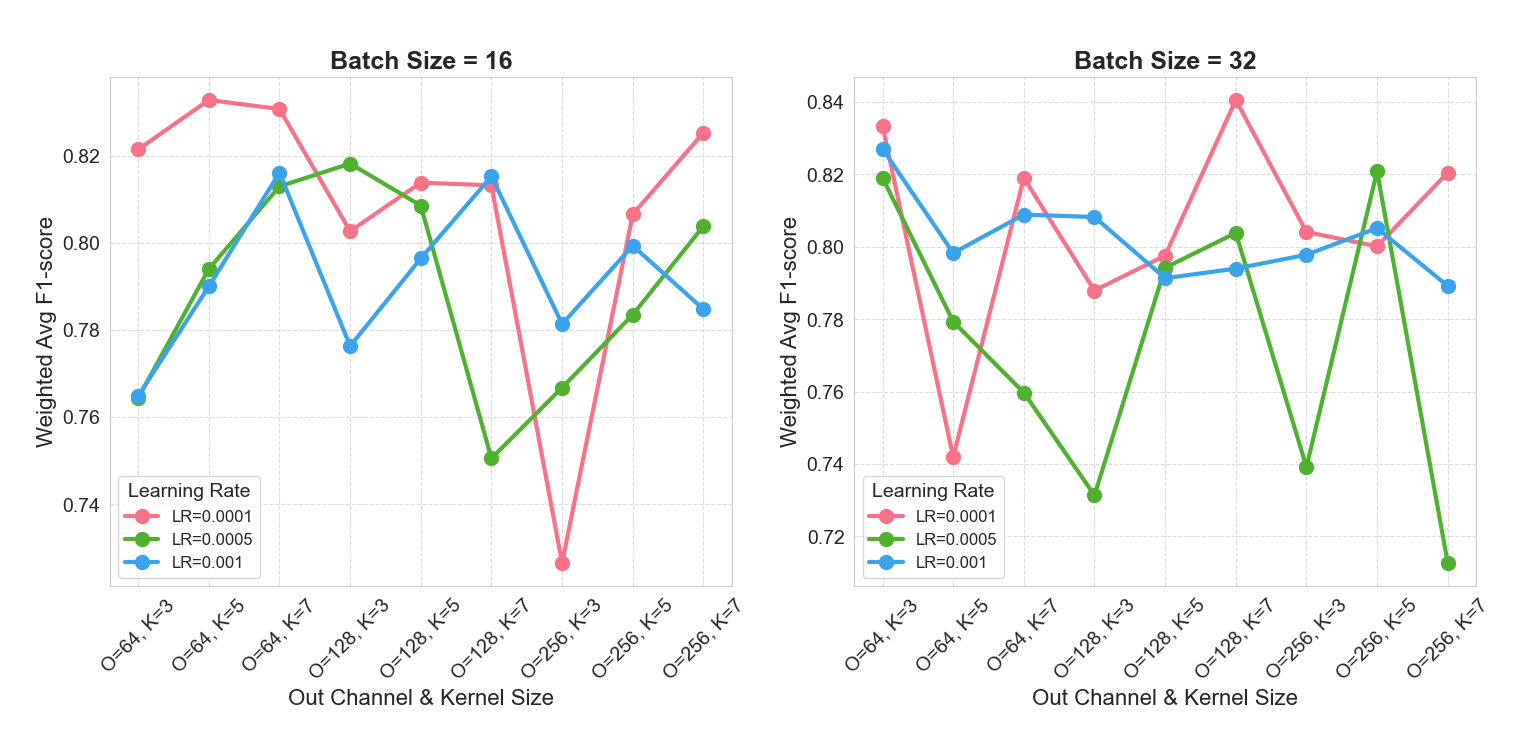}
    \caption{F1-Score Trends for CNN}
    \label{fig:fullwidth}
\end{figure}

The experiments tested CNN models with different hyperparameters for a binary classification problem. 
The best model, yielded an \textbf{accuracy of 84.00\%} and a \textbf{weighted F1-score of 84.05\%} using parameters such as 
128 out\_channels, kernel\_size=7, \texttt{dropout\_rate=0.3}, \texttt{learning\_rate=0.0001}, and \texttt{batch\_size=32}. 
Analysis indicated that bigger kernel sizes (e.g., 7 vs. 3/5) enhanced feature extraction by detecting larger contextual patterns, 
while moderate-sized \texttt{out\_channels} (128) balanced model complexity and generalization better than smaller (64) or bigger (256) sizes. Lower learning rates (\texttt{0.0001}) with dropout =0.3 increased stability and minimized overfitting, especially in deeper architectures. 
Interestingly, batch\_size=32 performed better than \texttt{batch\_size=16} in top models, indicating better gradient generalization. 
Increasing \texttt{out\_channels} to 256 sometimes enhanced performance (e.g., \texttt{F1=82.53\%}), but at the expense of potential overfitting, 
as evidenced by irregular performance across kernel sizes. Models with \texttt{kernel\_size=5} also competed (e.g., 
\texttt{F1=83.27\%}), but kernel\_size=7 consistently excelled, demonstrating the effectiveness of this setting for contextual learning. 
These results reinforce the need to combine kernel size, channel depth, and learning dynamics, with the best setup operating on the cross-road of receptive field breadth, regularization, and training stability.

\subsubsection{Result on LSTM}

\begin{figure}[H]
    \centering
    \includegraphics[width=\linewidth]{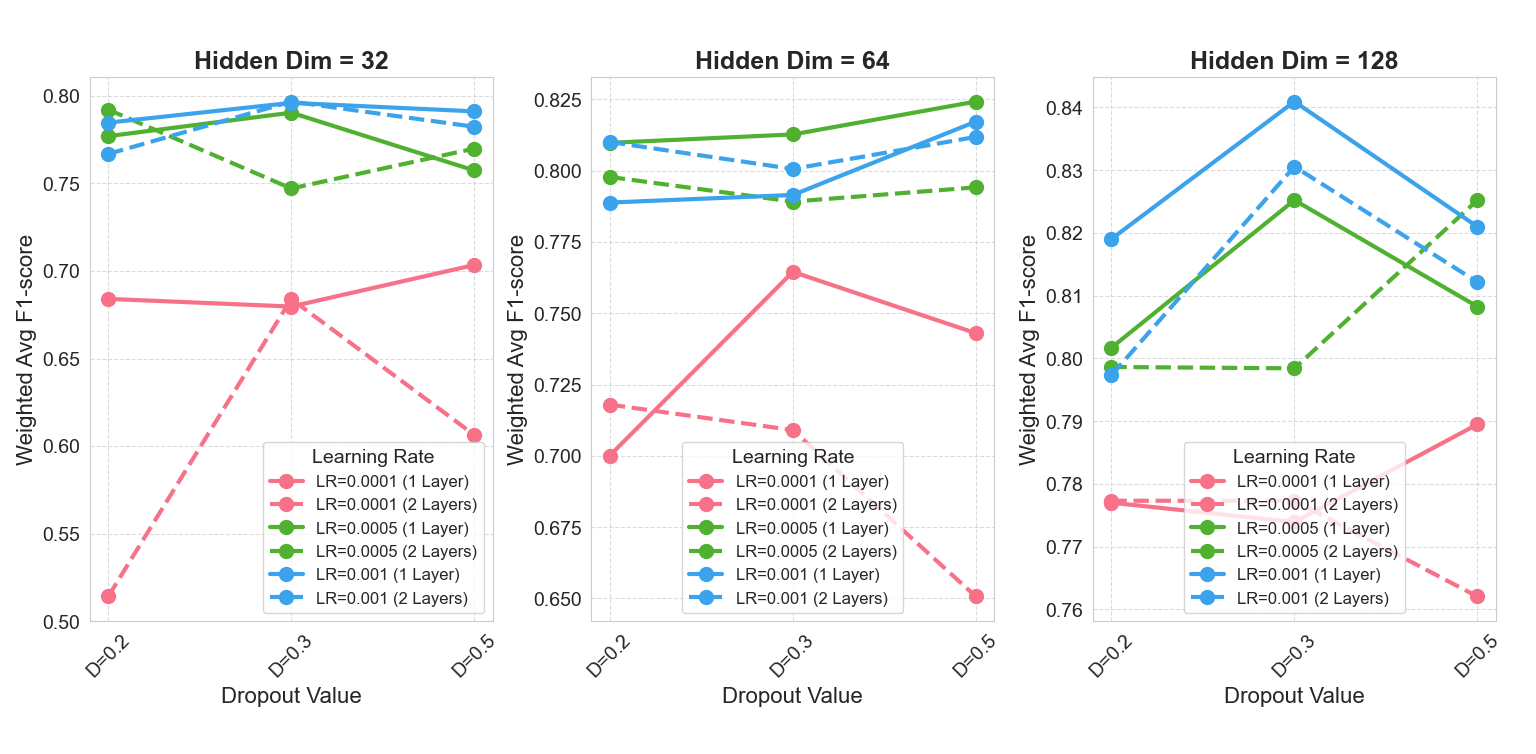}
    \caption{F1-Score Trends for LSTM}
    \label{fig:fullwidth}
\end{figure}

The tests compared LSTM models with different hyperparameters to select the best one for a binary classification problem. 
The top model, had an \textbf{accuracy of 84.25\%} and a \textbf{weighted F1-score of 84.09\%}, 
which far exceeded other configurations. Its configuration is a hidden dimension of 128, a single-layer unidirectional architecture, 
dropout of 0.3, learning rate of 0.001, and batch size of 16. 
Analysis showed that increased hidden dimensions (128 compared to 32/64) enhanced model capacity, allowing more detailed feature extraction without overfitting, 
especially in conjunction with moderate dropout (0.3). Bidirectional architectures were below par compared to unidirectional models. Bidirectional architectures reached 83.06\% F1 compared to 84.09\% for the unidirectional version), indicating unidirectional context was adequate for the task. 
Increased learning rates (0.001) with regularization (dropout=0.3) improved convergence, whereas smaller batch sizes (16 vs. 32) probably helped to increase gradient accuracy during training. 
Models with too much dropout (0.5) or reduced learning rates (0.0001) showed lower F1-scores from underfitting or slow convergence. 
Surprisingly, deeper architectures (2 layers) did not always beat single-layer models, suggesting diminishing returns from increased complexity. 
These results underscore the necessity to balance model capacity, regularization, and training dynamics, where the best combination accomplishes robust generalization 
by virtue of a synergistic interaction among hyperparameters.

\subsubsection{Result on LLM}
The comparative model analysis with different hyperparameters (learning rate and epochs) for a binary classification problem (\texttt{num\_labels=2}) demonstrates different performance patterns. 
Models trained with a learning rate of \texttt{1e-5} demonstrated incremental increases in accuracy as epochs were increased: 
76.25\% (2 epochs), 76.75\% (3 epochs), and 77.25\% (4 epochs), indicating stable but incremental learning. 
Still, the highest accuracy (\textbf{80.50\%}) was obtained by the model learning with a learning rate of \texttt{3e-5} and 3 epochs, 
suggesting the best balance between the speed of learning and the training time. 
It should be noted that the increase in epochs beyond 3 for this learning rate resulted in the decrease of accuracy (74.25\% at 4 epochs), presumably because of overfitting. 
On the other hand, higher learning rate models (\texttt{5e-5}) always performed poorly (57.75\% accuracy for all epochs), pointing towards the destabilizing influence of high learning rates. 
The top-performing model with (learning rate=\texttt{3e-5}, epochs=3), yielded an accuracy of \textbf{80.50\%}, 
showing that sensible learning rates coupled with careful epoch choice are paramount in achieving optimal performance in this task.
The results outlined above highlight key differences among the evaluated models, which are further examined and contextualized in the subsequent discussion.

\section{Discussion}\label{sec:discussion}
Through paired t-tests and performance measures, we compare the efficacy of conventional, deep learning, and large language models (LLMs) before scrutinizing crucial threats to validity that can affect the interpretation of findings.
This part offers an in-depth assessment of model performance on Nepali and Telugu code-mixed datasets by integrating statistical analysis, comparative measures, and an understanding of the relative strength and weaknesses of each methodology.

\subsection{Discussion on Nepali Dataset}

\begin{table}[h]
\caption{Pairwise t-Test with p-values for Model Comparison on Nepali Dataset}
\label{tab:model_comparison}
\centering
\begin{tabular}{@{}lccccccc@{}}
\toprule
 & \textbf{LR} & \textbf{SVM} & \textbf{RF} & \textbf{NN} & \textbf{CNN} & \textbf{LSTM} & \textbf{LLM} \\
\midrule
\textbf{LR}         & ---      & $<$ (0.975) & $>$ (0.001) & $>$ (0.002) & $>$ (0.078) & $>$ (0.257) & $<$ (0.001) \\
\textbf{SVM}        & $>$ (0.975) & ---      & $>$ (0.003) & $>$ (0.000) & $>$ (0.045) & $>$ (0.093) & $<$ (0.001) \\
\textbf{RF}         & $<$ (0.001) & $<$ (0.003) & ---      & $>$ (0.369) & $<$ (0.687) & $<$ (0.009) & $<$ (0.001) \\
\textbf{NN}         & $<$ (0.002) & $<$ (0.000) & $<$ (0.369) & ---      & $<$ (0.373) & $<$ (0.001) & $<$ (0.001) \\
\textbf{CNN}        & $<$ (0.078) & $<$ (0.045) & $>$ (0.687) & $>$ (0.373) & ---      & $<$ (0.107) & $<$ (0.001) \\
\textbf{LSTM}       & $<$ (0.257) & $<$ (0.093) & $>$ (0.009) & $>$ (0.001) & $>$ (0.107) & ---      & $<$ (0.001) \\
\textbf{LLM}        & $>$ (0.001) & $>$ (0.001) & $>$ (0.001) & $>$ (0.001) & $>$ (0.001) & $>$ (0.001) & ---      \\
\bottomrule
\end{tabular}
\end{table}

The statistical analysis of model performance provides several key findings. \textbf{The LLM showed undisputed dominance over all other models}, displaying statistically significant differences (p $<$ 0.0001) in paired comparisons with conventional machine learning (e.g., Logistic Regression, SVM) and deep learning models (CNN, LSTM). This emphasizes the revolutionary potential of large language models for sentiment analysis tasks. Among classical models, \textbf{Random Forest performed better than Logistic Regression} (p=0.001) and \textbf{Neural Networks performed better than Logistic Regression} (p=0.002), although SVM had no significant benefit over Logistic Regression (p=0.975), indicating minimal value of linear kernel-based methods for this task. Interestingly, \textbf{SVM performed better than Random Forest} (p=0.003), indicating subtle performance differences among ensemble methods. In deep learning comparisons, \textbf{LSTMs had an edge over Random Forests} (p=0.009) and \textbf{Neural Networks} (p=0.0007), but \textbf{CNNs did not have a significant edge over LSTMs} (p=0.107), suggesting context-dependent effectiveness of recurrent over convolutional architectures.

Surprisingly, \textbf{CNNs had no significant advantage over Random Forests} (p=0.687), contradicting assumptions regarding automatic superiority of deep learning for embedded text representations. The \textbf{consistent superiority of the LLM} (p $<$ 0.0001 in all comparisons) is consistent with its ability to draw on contextual patterns of semantics, surpassing both feature-engineered conventional models and task-specialized neural designs. The findings highlight the need to match model type to data complexity, with the LLM becoming a model for high-stakes sentiment analysis tasks.

\subsection{Discussion on Telugu Dataset}
\begin{table}[ht]
\centering
\caption{Pairwise t-Test Results for Model Comparison on Telugu-English Dataset}
\label{tab:telugu_model_comparison}
\begin{tabular}{@{}lccccccc@{}}
\toprule
 & \textbf{LR} & \textbf{SVM} & \textbf{RF} & \textbf{NN} & \textbf{CNN} & \textbf{LSTM} & \textbf{LLM} \\
\midrule
\textbf{LR}    & ---      & $>$ (0.022) & $>$ (0.006) & $>$ (0.721) & $>$ (0.142) & $<$ (0.910) & $<$ (0.001) \\
\textbf{SVM}   & $<$ (0.022) & ---      & $>$ (0.462) & $<$ (0.070) & $<$ (0.588) & $<$ (0.022) & $<$ (0.001) \\
\textbf{RF}    & $<$ (0.006) & $<$ (0.462) & ---      & $<$ (0.031) & $<$ (0.419) & $<$ (0.012) & $<$ (0.001) \\
\textbf{NN}    & $<$ (0.721) & $>$ (0.070) & $>$ (0.031) & ---      & $>$ (0.381) & $<$ (0.633) & $<$ (0.001) \\
\textbf{CNN}   & $<$ (0.142) & $>$ (0.588) & $>$ (0.419) & $<$ (0.381) & ---      & $<$ (0.103) & $<$ (0.001) \\
\textbf{LSTM}  & $>$ (0.910) & $>$ (0.022) & $>$ (0.012) & $>$ (0.633) & $>$ (0.103) & ---      & $<$ (0.001) \\
\textbf{LLM}   & $>$ (0.001) & $>$ (0.001) & $>$ (0.001) & $>$ (0.001) & $>$ (0.001) & $>$ (0.001) & ---      \\
\bottomrule
\end{tabular}
\end{table}

The statistical contrast of model performances through paired t-tests ($\alpha=0.05$) following 10-fold cross-validation indicates clear hierarchical trends. The \textbf{Large Language Model (LLM)} showed unambiguous superiority by being statistically significantly better than all other models (p$\approx$0.0000 for all contrasts). The \textbf{LSTM} came in second, beating Logistic Regression (p=0.9097), Random Forest (p=0.0219), Neural Network (p=0.0118), and CNN (p=0.6327), although its superiority over the CNN was not statistically significant (p=0.6327). Standard models such as \textbf{Logistic Regression} and \textbf{Random Forest} always lagged behind, with Random Forest being significantly worse compared to SVM (p=0.0060) and LSTM (p=0.0219). Although the \textbf{SVM} performed better than Logistic Regression (p=0.0220) and Random Forest (p=0.0060), it could not outperform advanced architectures such as LSTM and LLM. Interestingly, the \textbf{CNN} and \textbf{Neural Network} had inconsistent performance, where there were no statistically significant differences in a number of pairwise comparisons (e.g., CNN vs. Neural Network, p=0.4194). These observations highlight the superior performance of sequence-based models (LLM, LSTM) over traditional methods in this task. The findings also point to the importance of epoch and learning rate tuning, indicated by the fact that the highest accuracy (80.50\%) was achieved using the optimal parameter setting of the LLM (lr=3e-5, epochs=3).

\begin{table}[ht]
\centering
\caption{Performance Comparison of Models on Nepali-English and Telugu-English Datasets}
\label{tab:performance}
\begin{tabular}{@{}lcccc@{}}
\toprule
\textbf{Model} & \textbf{Precision} & \textbf{Recall} & \textbf{F1 Score} & \textbf{Accuracy} \\
\midrule
\multicolumn{5}{@{}l}{\textbf{Nepali-English Dataset}} \\
Logistic Regression & 0.720 & 0.720 & 0.723 & 0.720 \\
SVM & 0.730 & 0.730 & 0.724 & 0.730 \\
Random Forest & 0.700 & 0.700 & 0.697 & 0.700 \\
Neural Network & 0.740 & 0.740 & 0.736 & 0.740 \\
CNN & 0.735 & 0.738 & 0.733 & 0.738 \\
LSTM & 0.738 & 0.733 & 0.729 & 0.733 \\
LLM & \textbf{0.790} & \textbf{0.788} & \textbf{0.786} & \textbf{0.788} \\
\addlinespace
\multicolumn{5}{@{}l}{\textbf{Telugu-English Dataset}} \\
Logistic Regression & 0.531 & 0.523 & 0.525 & 0.523 \\
SVM & 0.548 & 0.584 & 0.480 & 0.584 \\
Random Forest & 0.534 & 0.583 & 0.477 & 0.583 \\
Neural Network & 0.530 & 0.542 & 0.505 & 0.542 \\
CNN & 0.522 & 0.524 & 0.483 & 0.524 \\
LSTM & 0.535 & 0.548 & 0.537 & 0.548 \\
LLM & \textbf{0.910} & \textbf{0.907} & \textbf{0.907} & \textbf{0.907} \\
\bottomrule
\end{tabular}
\end{table}

Table 5 shows a comparative evaluation of different machine learning models on Nepali and Telugu datasets with respect to four performance measures: Precision, Recall, F1 Score, and Accuracy. In both datasets, the Large Language Model (LLM) by far surpasses the conventional and deep learning models with the highest score in all measures. To be precise, on the Nepali dataset, LLM achieves an F1 Score of 0.7858 and Accuracy of 0.7878, which is better than CNN and LSTM. The benefit is even more marked on the Telugu dataset, where LLM achieves an F1 Score of 0.9069 and Accuracy of 0.9070, reflecting a wide performance gap compared to traditional models such as SVM and Random Forest. The results emphasize the better generalization and language flexibility of LLMs for multilingual text classification tasks.

\subsection{Threats to Validity}

In conducting this study on abusive language detection within code-mixed Nepali and Telugu social media posts, several potential threats to validity were identified and are outlined below:

\textbf{Internal Validity:} The selection of comments from specific videos may introduce selection bias, potentially affecting the representativeness of the dataset. Additionally, relying on volunteer annotators could lead to annotation bias due to varying interpretations of what constitutes abusive content.

\textbf{External Validity:} The findings may have limited generalization beyond the specific linguistic and cultural contexts of Nepali and Telugu code-mixed social media. The dynamic nature of language and evolving social norms also pose challenges to the temporal validity of the developed models.

\textbf{Construct Validity:} Defining "abusive" language is inherently subjective, which may lead to inconsistencies in annotation. The complexity of code-mixed language further complicates the accurate identification and labeling of abusive content.

\textbf{Statistical Conclusion Validity:} Potential issues such as overfitting, especially when using complex models on limited data, could affect the reliability of the conclusions drawn. Ensuring appropriate statistical methods and validation techniques is crucial to mitigate this threat.

\textbf{Ethical Considerations:} Exposure of annotators to potentially harmful content raises concerns about their well-being. Additionally, collecting and using social media comments necessitates careful attention to privacy and ethical standards.

Acknowledging these threats is essential for interpreting the study's findings and understanding the limitations inherent in the research design. To highlight the novelty and relevance of our findings, we now review related work in this domain.

\section{Related Work}\label{sec:related_work}

The rise of code-mixed language as a dominant form of online communication, particularly on social media, has spurred increasing research interest in multilingual Natural Language Processing (NLP). However, most of the existing work has centered on high-resource language pairs such as Hindi-English, Spanish-English, and Arabic-English, while low-resource combinations like Telugu-English and Nepali-English remain significantly underexplored. This gap is particularly evident in the domain of abusive language detection, where the informal, context-rich, and often culturally nuanced nature of code-mixed language presents unique challenges.

Researchers have acknowledged the importance of building annotated datasets to advance code-mixed NLP. A prominent contribution is the DravidianCodeMix dataset, which includes Tamil-English, Malayalam-English, and Kannada-English comments annotated for sentiment and offensive language detection tasks \cite{chakravarthi2022dravidiancodemix}. This dataset has been used in shared tasks that aim to benchmark models on South Asian languages. Similarly, Khanuja et al. introduced a Hindi-English dataset for natural language inference using Bollywood dialogues, offering insights into reasoning and semantics in code-switched settings \cite{khanuja2020new}. However, such efforts rarely extend to low-resource pairs like Telugu-English or Nepali-English.

Datasets like HASOC \cite{mandl2019overview} and OffensEval \cite{zampieri2019predicting} focus on hate speech and offensive language but are mostly confined to Hindi-English or Urdu-English combinations. These datasets form the foundation for supervised learning benchmarks, yet do not account for the linguistic and cultural characteristics of other underrepresented South Asian languages. While synthetic code-mixed data generation using large language models has been explored \cite{zeng2022synthetic}, such methods often lack the realism and variation of actual user-generated content, making human-annotated corpora indispensable.

In terms of methodology, early approaches to abusive language detection in code-mixed text employed traditional machine learning models such as SVMs, Logistic Regression, and Random Forests \cite{waseem2016hateful,founta2018large}. These models relied on engineered features like n-grams, character-level cues, and part-of-speech tags. Although these techniques are interpretable and lightweight, they generally fall short in capturing deeper contextual or sequential information—especially in noisy or highly informal code-mixed settings.

The advent of deep learning has led to the adoption of neural models such as CNNs and LSTMs, which are more capable of handling the sequential and contextual complexity of code-mixed language \cite{perezrosas2018automatic}. Transformer-based models like multilingual BERT (mBERT) \cite{devlin2019bert}, XLM-R \cite{conneau2020unsupervised}, and IndicBERT \cite{kakwani2020indicnlpsuite} have further advanced the field by enabling cross-lingual understanding. However, the performance of these models is often constrained by the lack of representation for languages like Telugu and Nepali in the pretraining data.

Linguistic challenges also remain significant. The absence of a standardized Romanization system introduces orthographic inconsistencies that complicate tokenization and preprocessing. Additionally, the fluidity of code-switching—occurring at phrase, clause, or word level—makes it harder to design robust NLP pipelines. Annotation itself is a complex task in this domain, as offensive language can be subtle, culturally specific, or sarcastic. Studies have shown that inconsistencies in labeling, especially in subjective tasks, can lead to notable drops in model accuracy \cite{waseem2016hateful}.

Despite notable advances, Telugu-English and Nepali-English remain largely absent from most mainstream resources, benchmarks, and shared tasks. Given that these combinations are frequently used by millions of users in emotionally or politically charged discussions, the need for focused research is clear. Moreover, existing datasets tend to focus on sentiment or emotion detection rather than abuse-specific tasks in organically occurring Romanized content.

To address this gap, the current study introduces two novel, manually annotated datasets for classifying abusive and non-abusive content in Telugu-English and Nepali-English tex \footnote{\url{https://github.com/Ranveer098/Code-Mix-and-Code-Switch}}. The data, collected from real-world social media platforms, covers a variety of domains such as politics, entertainment, and daily conversation. By including language tagging, part-of-speech labels, and code-mixing statistics, this dataset serves as a valuable linguistic and computational resource for future multilingual NLP research.

\section{Conclusion}\label{sec:conclusion}
This research presents a significant step toward addressing the scarcity of resources for abusive language detection in low-resource, code-mixed language contexts. By curating and annotating a large dataset of Telugu-English and Nepali-English comments, we contribute a valuable linguistic asset that mirrors the actual language used in multilingual digital spaces. The dataset captures the complexity of informal online discourse, including transliteration inconsistencies, cultural nuances, and diverse syntactic patterns, all of which pose challenges for automated systems.

Through systematic experimentation, we assessed the performance of multiple machine learning, deep learning, and large language models. The findings revealed that while classical models like Logistic Regression, SVM, and Random Forest serve as reasonable baselines, they struggle with the intricacies of code-mixed text. Deep learning architectures such as LSTM and CNN performed better, particularly in handling sequential and contextual dependencies. However, the most notable performance gains came from transformer-based LLMs, which demonstrated robust generalization and significantly outperformed other methods across both language pairs.

In addition to performance analysis, we applied rigorous statistical tests, including paired t-tests, to validate the significance of model differences. These results affirm the reliability of our evaluation and underscore the importance of hyperparameter tuning and architecture selection in achieving optimal results for code-mixed abusive content detection.

Despite the promising outcomes, the work also highlights persistent challenges, such as inconsistent Romanization, limited domain coverage, and the subjective nature of abuse annotation. These issues suggest avenues for future improvement, including the integration of more refined preprocessing techniques, culturally sensitive annotation guidelines, and the inclusion of additional South Asian code-mixed languages.

Overall, this study lays a strong foundation for future research in code-mixed NLP by offering a benchmark dataset, performance insights, and practical considerations for model development. As multilingual communication continues to evolve online, the need for effective moderation tools that account for linguistic diversity becomes increasingly urgent. We hope this work will encourage further advancements in ethical and inclusive AI systems tailored to underrepresented language communities.

\section*{Declarations}
\begin{itemize}
\item Conflict of interest: The authors declare that they have no known competing financial interests or personal relationships that could have appeared to influence the work reported in this paper.
\end{itemize}

\bibliography{sn-bibliography}

\end{document}